\documentclass[sigconf]{acmart}

\usepackage{subcaption}
\usepackage{booktabs}
\usepackage{multirow}
\usepackage{graphicx}
\usepackage{adjustbox}
\usepackage{tabularx}
\usepackage{tcolorbox}
\usepackage{cleveref}
\usepackage{enumitem}
\usepackage{caption}

\AtBeginDocument{%
  }

\copyrightyear{2026}
\acmYear{2026}
\setcopyright{cc}
\setcctype{by}
\acmConference[KDD '26]{Proceedings of the 32nd ACM SIGKDD Conference on Knowledge Discovery and Data Mining V.2}{August 09--13, 2026}{Jeju Island, Republic of Korea}
\acmBooktitle{Proceedings of the 32nd ACM SIGKDD Conference on Knowledge Discovery and Data Mining V.2 (KDD '26), August 09--13, 2026, Jeju Island, Republic of Korea}
\acmDOI{10.1145/3770855.3817994}
\acmISBN{979-8-4007-2259-2/2026/08}

\begin{document}


\title{Can Structural Cues Save LLMs?\\ Evaluating Language Models in Massive Document Streams}

\author{Yukyung Lee}
\affiliation{%
  \institution{Boston University}
  \city{Boston}
  \country{USA}}
\email{ylee5@bu.edu}

\author{Yebin Lim}
\authornote{equal contribution}
\affiliation{%
  \institution{Korea University}
  \city{Seoul}
  \country{Korea}}
\email{yebinuni@korea.ac.kr}

\author{Woojun Jung}
\authornotemark[1]
\affiliation{%
  \institution{Korea University}
  \city{Seoul}
  \country{Korea}}
\email{woojoon@korea.ac.kr}

\author{Wonjun Choi}
\affiliation{%
  \institution{Korea University}
  \city{Seoul}
  \country{Korea}}
\email{migreeni@korea.ac.kr}

\author{Susik Yoon}
\affiliation{%
  \institution{Korea University}
  \city{Seoul}
  \country{Korea}}
\email{susik@korea.ac.kr}
\renewcommand{\shortauthors}{Lee et al.}

\begin{abstract}
Evaluating language models in streaming environments is critical, yet underexplored. Existing benchmarks either focus on single complex events or provide curated inputs for each query, and do not evaluate models under the conflicts that arise when multiple concurrent events are mixed within the same document stream. We introduce StreamBench, a benchmark built from major news stories in 2016 and 2025, comprising 605 events and 15,354 documents across three tasks: Topic Clustering, Temporal Question Answering, and Summarization. To diagnose how models fail, we compare performance with and without structural cues, which organize key facts by event. We find that structural cues improve performance on clustering (up to +4.37\%) and temporal QA (up to +9.63\%), helping models locate relevant information and separate distinct events. While temporal reasoning remains an open challenge inherent to current LLMs, consistent gains across tasks show that structural cues are a promising direction for future work in massive document streams.
\end{abstract}




\maketitle
\newcommand\kddavailabilityurl{https://doi.org/10.5281/zenodo.20501891}
\ifdefempty{\kddavailabilityurl}{}{
\begingroup\small\noindent\raggedright\textbf{Resource Availability:}\\
The source code of this paper has been made publicly available at \url{\kddavailabilityurl}, with the source code repository at \url{https://github.com/yukyunglee/llm-streaming-eval}.
\endgroup
}

\section{Introduction}
\label{sec:intro}

\begin{figure}[t]
    \centering
    \includegraphics[width=\columnwidth]{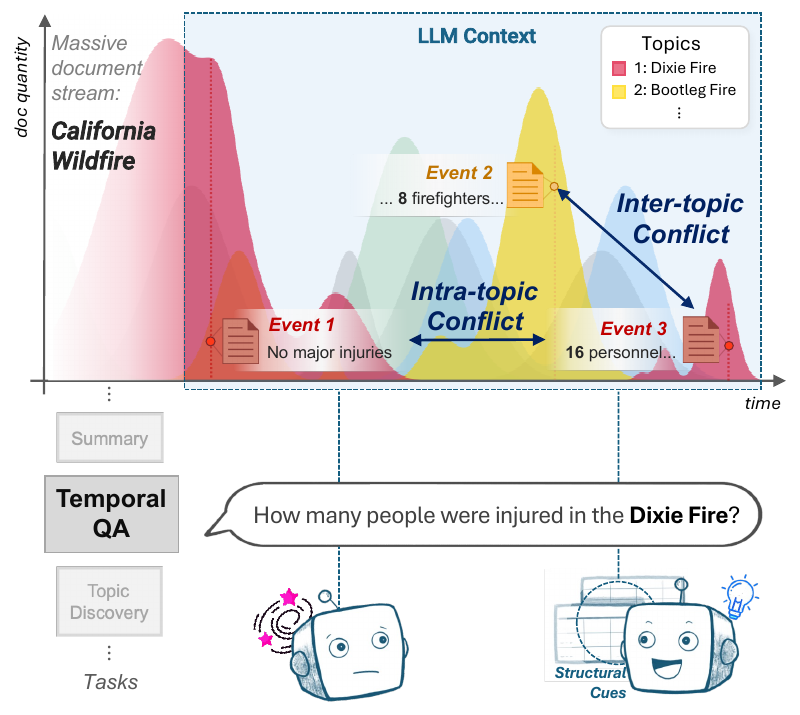}
    \caption{Two challenges in streaming environments. Intra-topic conflict: within Dixie Fire, the most recent event (Event 3) has fewer documents than the outdated event (Event 1), making it harder to find the latest information. Inter-topic conflict: when asked about Dixie Fire (Topic 1), the LLM may confuse ``8 firefighters'' (Event 2) from Bootleg Fire (Topic 2).}
    \label{fig:motivating}
\end{figure}

Information streams evolve in real time, particularly in domains such as news, where documents continuously arrive from diverse sources and facts are updated over time. A single news story often spans multiple topics and associated events, where the corresponding documents emerge intertwined within the same stream.
This document stream setting gives rise to a range of downstream tasks, including topic discovery~\citep{10.5555/772260.772262, 10.1145/3543507.3583507, 10.1145/3269206.3269309, nakshatri-etal-2023-using}, question answering (QA)~\citep{zhang-choi-2021-situatedqa, streamingqa2022}, and summarization~\citep{10.1145/3543507.3583371, song-etal-2025-temporal}. 
Traditionally, each of these tasks has been addressed by task-specific models developed in an ad-hoc manner~\citep{10.1145/3373464.3373470, 10.1145/3704922}. With the rise of foundation models, there is growing interest in applying large language models (LLMs) directly to document streams, leveraging their broad capabilities to handle multiple tasks with a single model~\citep{vu-etal-2024-freshllms, dai2025are}. However, LLMs operate within a fixed context window. As new documents arrive, old information and new updates share the same window, making it increasingly difficult to identify what is relevant and up to date~\citep{du-etal-2025-context}. Such inherently dynamic and unbounded streaming settings pose nontrivial challenges for LLMs, beyond what a typical static, long-context setting presents.

Specifically, we identify two key conflicts. First, intra-topic conflict: within a single topic, documents accumulate over time as new events occur, biasing models toward older information and making it harder to identify the most recent state. Second, inter-topic conflict: when documents from multiple related topics overlap in the same context window, models struggle to distinguish which facts belong to which topic. These conflicts place a significant cognitive burden on models, as they must organize scattered information by topic and time while reasoning over them~\citep{levy-etal-2024-task, li2025longcontext}. 
In~\Cref{fig:motivating}, for example, a news story about California Wildfire contains multiple topics, such as Dixie Fire and Bootleg Fire, where fresh news articles from associated events continuously feed into a document stream. When an LLM is asked to reason in the latest context of the stream (e.g., "How many people were injured in the Dixie Fire?"), it faces difficulty identifying exact information in Event 3, which is more recent but has been reported by fewer articles than Event 1 (i.e., intra-topic conflict). The confusion intensifies when the model needs to distinguish relevant information from concurrent topics, such as "8 firefighters" from Bootleg Fire (i.e., inter-topic conflict).  

However, existing benchmarks fall short in evaluating LLM capabilities in such streaming settings. While several time-sensitive datasets have been proposed~\citep{zhang-choi-2021-situatedqa, streamingqa2022, kasai2023realtime, vu-etal-2024-freshllms}, they primarily evaluate temporally grounded questions over static snapshots, rather than continuously expanding contexts with accumulating information. Other recent benchmarks address complex event understanding~\citep{zhang-etal-2024-analyzing}, but focus on single events, failing to capture scenarios where multiple events develop concurrently with varying quantities. Moreover, existing evaluations report only end-to-end performance without diagnosing \textit{why} models fail, and focus on identifying the problem rather than exploring directions to address it. 

In this work, we present a systematic and comprehensive evaluation of LLM capabilities in streaming environments, featuring the following contributions:
\begin{itemize}[leftmargin=10pt, noitemsep]
\item First, we construct \textbf{StreamBench}, a benchmark built from real-world news streams spanning 2016 and 2025, comprising 605 events and 15,354 documents. We evaluate seven LLMs of varying scales (1B--123B) across three tasks: Topic clustering, Temporal QA, and Summarization. These tasks capture distinct cognitive demands of streaming environments: organizing scattered information, answering specific questions, and compressing accumulated texts over evolving stories. StreamBench features dynamic concurrent events with varying text quantity, capturing the core characteristics of streaming environments. 

\item Second, to diagnose why models struggle under these settings, we introduce \textbf{structural cues} as a diagnostic probe. The core difficulty in real-world streaming environments is that information from multiple topics arrives mixed together over time, leading to intra- and inter-topic conflicts. Prior work in a static setting has shown that structured knowledge representations help models handle complex information~\citep{10.1109/TKDE.2024.3352100, wu2024thinking}.
If failures primarily stem from difficulties in organizing scattered information, providing auxiliary organizational support should alleviate these issues. Based on this intuition, we design the simplest form of structural cues as key facts and entities organized by topic, serving as supplementary signals for LLMs. By comparing performance with and without these cues, we identify what aspects become easier with organization, and what difficulties remain.
\end{itemize}

Our evaluation shows that current LLMs struggle in streaming environments across all three tasks. Comparing performance with and without structural cues, we observe that they provide partial improvements, helping models \emph{find} relevant information. Structural cues improve event separation in clustering (up to 4.37\% in B$^3$ F1) and temporal ordering in QA (up to 9.63\% in accuracy). However, reasoning over the located information remains challenging. In clustering, precise boundary detection remains difficult despite reduced over-clustering. In temporal QA, models still struggle with tracking the current state of entities even when information is well-organized. In summarization, structural cues show smaller gains (up to 0.87\% in ROUGE-L and 3.40\% in METEOR).
Overall, structural cues consistently help models find and organize information, while reasoning over temporal dynamics remains an open challenge inherent to current LLMs. We believe our findings and benchmark can motivate further exploration into how LLMs handle conflicts in streaming document environments.

\section{Related Work}
\label{sec:rel-work}

\subsection{Temporal and Streaming Evaluation}

Recent benchmarks evaluate how LLMs handle time-varying knowledge. Streaming QA \citep{streamingqa2022}, RealTime QA \citep{kasai2023realtime}, and FreshQA \citep{vu-etal-2024-freshllms} test whether models rely on up-to-date information rather than outdated parametric knowledge. Daily Oracle \citep{dai2025are} evaluates this ability over continuously updated news, enabling longitudinal analysis of how performance degrades as information becomes stale. More recently, HoH \citep{ouyang-etal-2025-hoh} finds that even when correct answers are present, outdated context significantly impairs LLM reasoning, underscoring the fragility of real-time processing.  

These benchmarks focus on temporal accuracy for individual topics or events. In contrast, real-world streams contain multiple concurrent events that evolve at different rates, while unrelated information accumulates together. In this work, we focus on streaming environments, where information arrives continuously and must be interpreted incrementally. We define two conflicts specific to streaming settings (intra-topic conflict and inter-topic conflict) and analyze their impact on performance.

\subsection{Event-Centric Document Understanding}

Tracking evolving topics and narratives across multiple documents is a core challenge in real-world streaming scenarios such as news. Prior work has addressed this through Topic Detection and Tracking  \citep{10.5555/772260.772262}, multi-document summarization \citep{10.1145/1008992.1009065}, and document-level event extraction \citep{xu-etal-2021-document}. Recent efforts apply LLMs to this setting: \citet{hu-etal-2024-moments} incrementally update event timelines as new articles arrive; PDSum \citep{10.1145/3543507.3583371} performs prototype-based continuous summarization; SCStory \citep{10.1145/3543507.3583507} uses self-supervised learning to track narrative evolution; and \citet{song-etal-2025-temporal} show that better temporal reasoning enhances summarization accuracy.

These methods aim to improve end-to-end output quality but offer limited interpretability into why models fail. We take a complementary diagnostic perspective, using structural cues as a probe to identify the sources of model failures. By comparing performance on raw streaming text versus inputs with explicit structural cues, we measure how much organization helps and what difficulties remain even with organized inputs.

\section{Problem Formulation}
\label{sec:formulation}

\subsection{Streaming Environment}

In real-world streaming environments, documents from multiple topics arrive mixed together over time. We define the following structure: a \textit{story} is a high-level news narrative (e.g., California Wildfire) that contains $M$ \textit{topics} $\{T_1, \ldots, T_M\}$ (e.g., Dixie Fire, Bootleg Fire), where $m \in \{1, \ldots, M\}$ indexes each topic. Each topic $T_m$ consists of $N_m$ events $\{e_{m,1}, \ldots, e_{m,N_m}\}$, where $n \in \{1, \ldots, N_m\}$ indexes events within topic $m$. An event $e_{m,n}$ is a temporally localized at timestamp $t_{m,n}$, described by $K_{m,n}$ documents $\{d_{m,n,1}, \ldots, d_{m,n,K_{m,n}}\}$, where $k \in \{1, \ldots, K_{m,n}\}$ indexes documents within event $e_{m,n}$. Each document inherits the timestamp of its event.

Documents arrive chronologically, regardless of topic:
\begin{equation}
X = \{ d_{m,n,k} \} \;\text{ordered by } t_{m,n}.
\end{equation}
LLMs observe only the latest set of documents and their timestamps, without topic or event labels.

\paragraph{Sliding Window.} We define $J$ sliding windows of $w$ days with a $s$-day stride, where $j \in \{1, \ldots, J\}$ indexes each window. 
Each window $W_j$ contains all documents within that period and serves as the model input at time step $j$:
\begin{equation}
W_j = \{ d_{m,n,k} \mid t_{\text{start}}^j \le t_{m,n} < t_{\text{start}}^j + w \}.
\end{equation}
In our experiments, we set $w=7$ and $s=1$.

\paragraph{Controlling Document Stream Volume.} We sample $k$ documents per event to control the volume of streams, e.g., $k \in \{1, 3, 5, 10\}$. Larger $k$ increases inter-topic conflict by mixing more documents from different topics, and intra-topic conflict by accumulating more documents for earlier events. 

\subsection{Evaluation Tasks}
We select three tasks that capture distinct demands of streaming environments. Topic clustering requires separating documents from different topics within the stream. Temporal QA requires locating relevant information and reasoning over its temporal order.  Summarization requires compressing information across multiple topics.

\subsubsection{Task 1: Topic Clustering}

Given a window $W_j$ containing documents from multiple topics, assign each document to a topic:
\begin{equation}
f_{\text{cluster}}: d_i \in W_j \rightarrow \hat{t}_i \in \{1, 2, \ldots\},
\end{equation}
where $\hat{t}_i$ is the predicted topic ID for document $d_i$. The model may assign documents to existing topics or create new topics as needed. Documents within the window are processed one at a time in chronological order. When the first document arrives, the model creates an initial topic and extracts representative keywords. As each subsequent document arrives, the model decides whether it belongs to an existing topic or represents a new topic, then updates the topic's keywords accordingly. We use B$^3$ F1, which calculates precision and recall for each document and averages them.

\subsubsection{Task 2: Temporal Question Answering}

Given a question $q$, multiple-choice options $\mathcal{A} = \{a, b, c, d\}$, and documents in window $W_j$, predict the correct answer:
\begin{equation}
f_{\text{QA}}: (q, \mathcal{A}, W_j) \rightarrow \hat{a} \in \mathcal{A}.
\end{equation}
The model must find relevant information from documents spanning multiple topics and derive the correct answer considering temporal order. Each question is annotated with a timestamp $t_q$. The model receives only documents up to that timestamp:
\begin{equation}
W_j^{(q)} = \{d_{m,n,k} \in W_j \mid t_{m,n} \leq t_q\}.
\end{equation}
This reflects realistic streaming situations where future information is inaccessible. When conflicting information exists across time points, the model must prioritize the most recent information. We use accuracy as the evaluation metric.

\subsubsection{Task 3: Summarization}

Given all documents in window $W_j$, generate a multi-topic summary:
\begin{equation}
f_{\text{summ}}: W_j \rightarrow \hat{y}.
\end{equation}
The model must identify multiple topics, extract key information from each, and integrate them into a coherent summary. Unlike single-topic summarization, it must maintain balanced coverage across all topics while preserving temporal consistency. We use ROUGE-L~\citep{lin2004rouge} as the primary metric, with BLEU~\citep{papineni2002bleu} and METEOR~\citep{banerjee-lavie-2005-meteor} as supplementary metrics.\footnote{
Although BERTScore~\citep{zhang2019bertscore} is widely used for summarization evaluation, in our preliminary experiments it showed high sensitivity to length differences between generated and reference summaries, often reflecting output length rather than content quality. \citet{fabbri-etal-2021-summeval} also report that BERTScore shows lower correlation with human judgments than ROUGE and METEOR in summarization evaluation.} We also employ LLM-as-a-judge for multi-dimensional verification. 

\subsection{Structural Cues}

To diagnose why models fail under streaming environment, we compare performance with and without structural cues.

\subsubsection{Cue Definition.} For each event $e_{m,n}$, the structural cue $s_{m,n}$ is defined as:
\begin{equation}
s_{m,n} = (\text{People}, \text{Location}, \text{Result}, \text{EventAttr}),
\end{equation}
where People and Location are lists of key entities, Result is a summary of the event's main outcome, and EventAttr includes attributes such as cause and effect. We extract cues using LLMs, followed by human verification. Structural cues do not add new information but reorganize existing information by event. To prevent the LLM from introducing its own knowledge during extraction, we strictly constrain cues to contain only terms that appear in the source documents.

We compare two conditions for the same window $W_j$:
\begin{itemize}[leftmargin=10pt, noitemsep]
    \item \textbf{Raw Input}: the model receives only documents in $W_j$, where documents from multiple topics are mixed together.
    \item \textbf{Cued Input}: the model receives documents in $W_j$ along with structural cues $\{s_{m,n}\}$ for all events in the window, reducing the burden of organization.
\end{itemize}

\subsubsection{Efficacy Quantification.} For each task's evaluation metric $\mathcal{M}$, we define:
\begin{equation}
\Delta_{\text{org}} = \mathcal{M}(\text{Cued}) - \mathcal{M}(\text{Raw}) \text{ and } \Delta_{\text{gap}} = \mathcal{M}_{\text{ceiling}} - \mathcal{M}(\text{Cued}),
\end{equation}
where $\mathcal{M}_{\text{ceiling}}$ is the theoretical upper bound (e.g., 100 for clustering B$^3$ F1 and QA accuracy, ROUGE-L, given all metrics are reported on a scale of 0 to 100 in this paper). $\Delta_{\text{org}}$ denotes the performance gains of the organization. $\Delta_{\text{gap}}$ represents the residual gap to the ceiling even after incorporating structural cues into the LLM context \footnote{We note that lexical-overlap metrics such as ROUGE-L are not expected to reach 100 in abstractive summarization, even for high-quality summaries. Accordingly, $\Delta_{\text{gap}}$ for summarization should be interpreted as a relative comparison rather than as an absolute measure of remaining difficulty.}.
\section{StreamBench}
\label{sec:streambench}

To apply our diagnostic framework, we require a dataset that enables: (1) realistic streaming environments with multiple concurrent events, (2) systematic control of document stream volume, and (3) construction of structural cues for each event. 

We construct StreamBench to diagnose LLM failures in streaming news environments. StreamBench comprises six news stories from two time periods: 2025 and 2016. The 2025 data captures recent events mostly occurring after most LLMs' knowledge cutoffs, minimizing parametric knowledge influence. The 2016 data allows for verifying the consistency of findings across periods. We selected stories with diverse temporal distributions; some spanned a full year, while others concentrated in specific periods. 

\begin{table}[t]
\caption{Dataset statistics. StreamBench comprises six news stories spanning two time periods: 2024–25 (Stories A–C) and 2016 (Stories D–F). Token counts are measured using the \texttt{Llama-3} tokenizer.}
\label{tab:dataset-statistics}
\centering
\renewcommand{\arraystretch}{1.4}
\resizebox{\columnwidth}{!}{%
\begin{tabular}{cllcrrrrr}
\toprule
Story & Contents & Duration & Topics & Events & Docs & Avg. Tok \\
\midrule
A & California Fire & Jan'24–Nov'25 & 32 & 113 & 768 & 1,555 \\
B & South Korea Martial Law & Jun'24–Nov'25 & 15 & 108 & 1,135 & 905 \\
C & 60th US Presidential Election & Jan'24–Nov'25 & 32 & 111 & 724 & 2,061 \\
\midrule
D & Summer Olympics & Apr'16–Dec'16 & 20 & 35 & 977 & 1,296 \\
E & Israel-Palestine Conflict& Jan'16–Dec'16 & 13 & 45 & 387 & 732 \\
F & 58th US Presidential Election & Jan'16–Dec'16 & 88 & 193 & 11,363 & 1,916 \\
\midrule
\multicolumn{3}{l}{Total} & 200 & 605 & 15,354 & -- \\
\bottomrule
\end{tabular}}
\end{table}

\subsection{Data Collection}

\paragraph{2025 Stories (A-C)} We collected three stories: California Wildfires (Story A), South Korea Martial Law (Story B), and 60th US presidential election (Story C). For each story, we extracted event structures from Wikipedia pages based on section headings (Background, Aftermath, Impact, Response). Related news articles were collected via NewsAPI\footnote{\url{https://newsapi.ai/}}. We extracted the top 5 most frequent named entities from each event's summary using spaCy, and when extracted keywords were insufficient, combined them with predefined fallback keywords (e.g., ``California AND (wildfire OR fire)'' for California Fire). Searches were restricted to English articles from the event occurrence date, collecting up to 30 articles per event ranked by relevance. After token-based deduplication, we used the Newspaper library to replace truncated content with full article text. We then computed cosine similarity with event summaries using Sentence Transformers~\citep{reimers-2019-sentence-bert} (\texttt{gte-large-en-v1.5}) and removed articles below a 0.6 threshold, yielding 768 (Story A), 1,135 (Story B), and 724 (Story C) articles.
\paragraph{2016 Stories (D--F)} We curated three stories from the W2E~\citep{10.1145/3269206.3269309} dataset: Summer Olympics (Story D), Israel-Palestine Conflict (Story E), and 58th US Presidential Election (Story F). W2E defines events based on Wikipedia and maps each event to a set of major news agency articles. \Cref{fig:time-distribution} shows each story's temporal distribution.

\begin{figure}[t]
    \centering
    \includegraphics[width=\columnwidth]{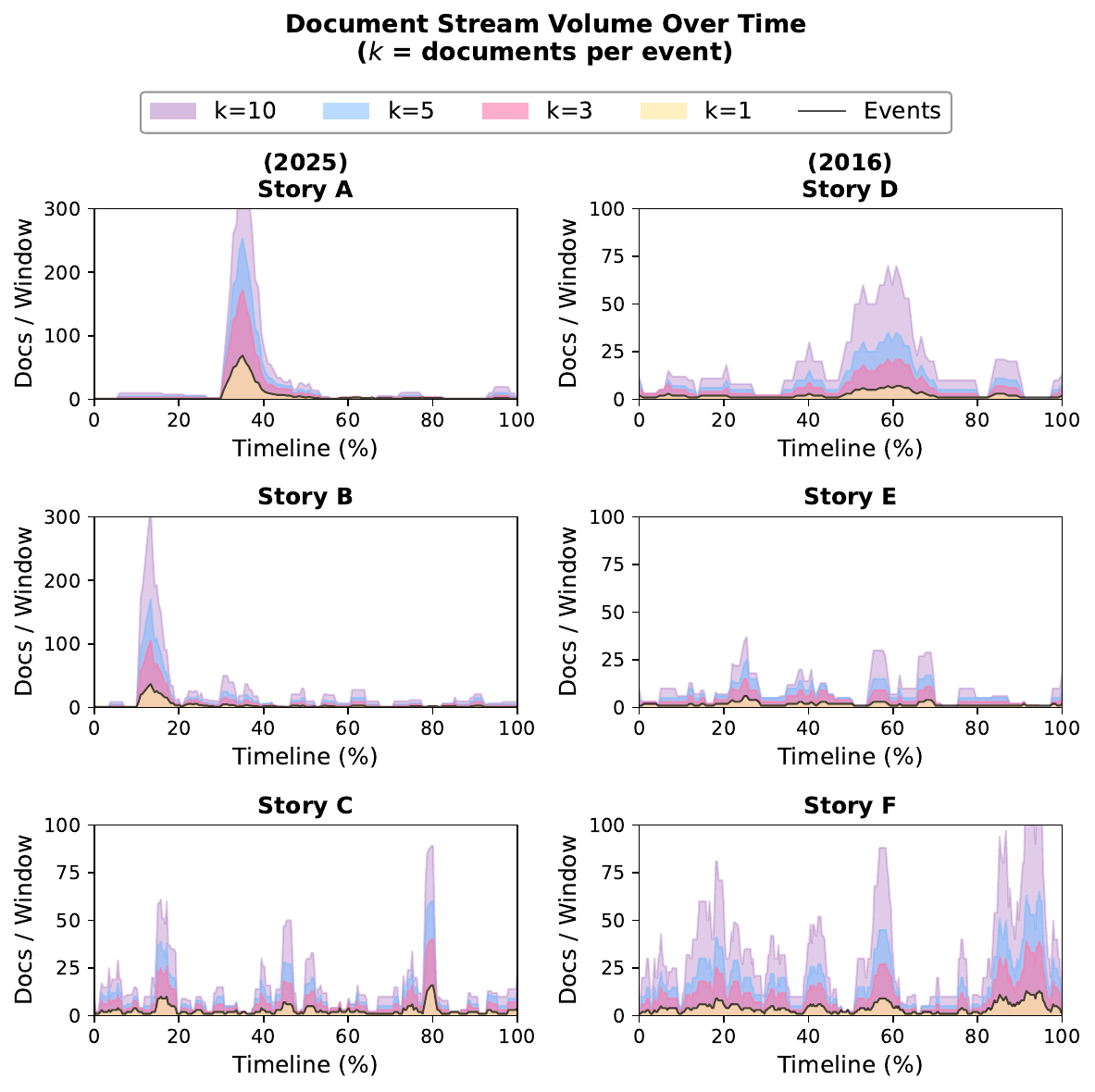}
    \caption{Document stream volume over time. The x-axis shows the normalized story timeline, and the y-axis indicates the number of documents per 7-day window. We vary $k \in {1, 3, 5, 10}$, the number of documents sampled per event. The 2016 stories (D–F) contain more documents per window than the 2025 stories (A–C).}
    \label{fig:input-density}
\end{figure}

\subsection{Dataset Statistics}
\Cref{tab:dataset-statistics} summarizes StreamBench statistics. The dataset comprises 200 topics, 605 events, and 15,354 documents. Story sizes and temporal distributions vary, with average tokens per document ranging from 732 (Story E) to 2,061 (Story C). StreamBench includes 1,087 QA pairs, 605 summarization annotations, and 200 clustering annotations. Due to the sliding window design (7-day window, 1-day stride), each annotation can appear across multiple consecutive windows, yielding window-level evaluation instances across 1,246 windows: 6,933 QA, 4,150 summarization, and 3,026 clustering instances. \Cref{fig:input-density} shows event distribution over time for each story. Story F (58th US Presidential Election) is distributed evenly over a year, while Story B (South Korea Martial Law) concentrates in a specific period. This diversity enables evaluating model behavior under different temporal patterns.

\begin{figure*}[t]
\centering
    \includegraphics[width=\textwidth]{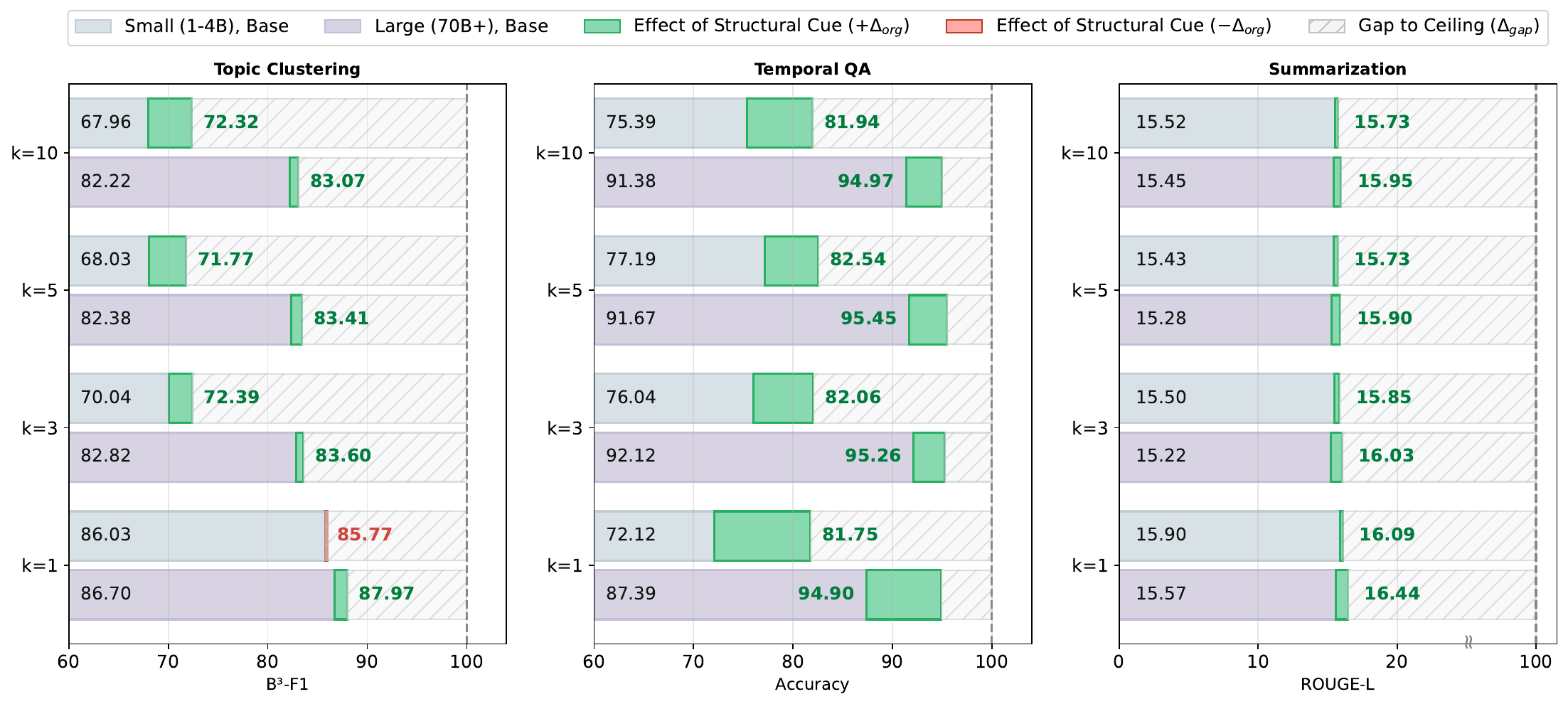}
    \caption{Performance bottleneck analysis across tasks and model scales. Stacked bar charts show base performance and structural cue effects for Small (1--4B) and Large (70B+) models. Green bar indicates positive effect from structural cues; red indicates negative effect. Hatched bar represents headroom to the ceiling. $k$ indicate number of documents sampled per event.}
\label{fig:bottleneck}
\end{figure*}

\subsection{Structural Cue Extraction}
Structural cues were constructed through a GPT-4o-based multi-stage extraction pipeline~\citep{achiam2023gpt4}. For each event, we selected the article with the highest similarity to the event summary, then performed three-stage extraction. Stage 1 extracts People, Location, and Involved Organizations. People include only explicitly mentioned individual names, excluding groups or titles alone. Stage 2 determines Event Type, Cause, Effect, Action, and Sentiment. Stage 3 identifies Result and Stakeholders, with Result classified into Policy, Impact, Action, Status, and Agenda. Each stage uses few-shot prompting and JSON schemas to ensure consistent formatting.

\subsection{Task-Specific Annotation}
\label{subsec:annotation}
\paragraph{Topic Clustering} Ground truth clusters are derived from topic labels. For 2016 data, we used topic labels provided by W2E dataset. For 2025 data, we defined initial topics based on Wikipedia structure, with remaining events assigned to the most similar topic based on cosine similarity.
\paragraph{Temporal QA} We generated multiple-choice questions for events with sufficient structural cues. Questions are classified into two types: Result Recognition requires reasoning about causal relationships between temporally separated events, while Entity Tracking requires tracking entity states over time and prioritizing recent information. QA generation jointly considered questions, answers, and options under strict constraints to ensure answer validity and distractor quality. After automatic validation, 1,087 pairs that satisfied all constraints were included in the final dataset out of the 1,483 initially generated. Answer distribution is A (26.1\%), B (27.7\%), C (23.3\%), D (22.9\%), showing no positional bias. To assess verification reliability, we further conduct human verification on a randomly sampled subset of 108 QA pairs (10\% of the set). Human verification showed 83.3\%, 80.6\%, 80.6\% agreement across three annotators (authors of this paper). Disagreements arose from borderline cases (lexical leakage, temporal assumptions, granularity mismatches) rather than factual errors. 
\paragraph{Summarization} Reference summaries were constructed from human-written event descriptions collected from the Wikipedia Event Portal\footnote{\url{https://en.wikipedia.org/wiki/Portal:Current_events}} and Wikipedia articles. Since each event has its own independently written description, we concatenated them in chronological order within each window and used GPT-4o to consolidate them into fluent, non-redundant multi-document summaries without altering the factual content.

\paragraph{Temporal QA} We generated multiple-choice questions for events with sufficient structural cues, classified into two types. \textit{Result Recognition} questions (e.g., ``What was the result of [event]?'') require reasoning about causal relationships between temporally separated events, while \textit{Entity Tracking} questions (e.g., ``Who/Where is currently [role/status]?'') require tracking entity states over time and prioritizing recent information. The two types account for 623 (57.3\%) and 464 (42.7\%) questions, respectively.

QA generation jointly considers questions, answers, and choices under strict constraints. Answers are selected from structured cue fields (e.g., Result, People, Location) and must be supported by the referenced articles, with specificity aligned to the reported information. During question generation, we avoid answer strings or lexically identical phrasing from the source, exclude static attributes (e.g., birthplace), and tie each question to a temporal reference. Choices are built in two stages: we generate a pool of 10 plausible candidates, then select a balanced subset of distractors that are mutually exclusive and consistent in specificity. After automatic validation with GPT-4o, 1{,}087 of the 1{,}483 generated pairs satisfied all constraints and were retained. Answer distribution is A (26.1\%), B (27.7\%), C (23.3\%), D (22.9\%), showing no positional bias. To assess reliability, three annotators (authors) independently verified a random 10\% subset (108 pairs), showing 80.6--83.3\% agreement with automatic validation; disagreements arose from borderline cases (lexical leakage, temporal assumptions, granularity mismatches) rather than factual errors.
\section{Experiment}

\begin{figure*}[ht]
    \centering
    \begin{subfigure}[b]{0.33\textwidth}
        \includegraphics[width=\textwidth]{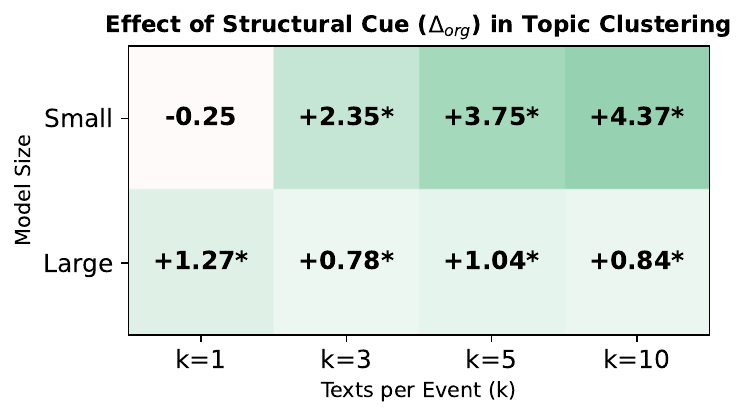} 
        \label{fig:subfig1}
        \vspace{-0.5cm}
        \caption{Topic Clustering.}
    \end{subfigure}
    \hfill
    \begin{subfigure}[b]{0.33\textwidth}
    \includegraphics[width=\textwidth]{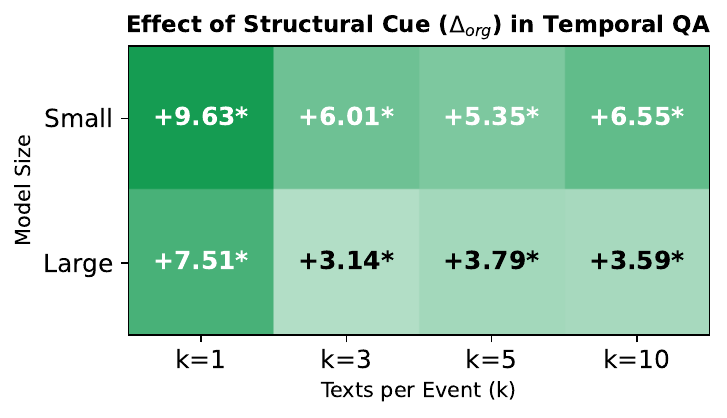}
        \label{fig:subfig2}
        \vspace{-0.5cm}
        \caption{Temporal QA.}
    \end{subfigure}
    \hfill
    \begin{subfigure}[b]{0.33\textwidth}
        \includegraphics[width=\textwidth]{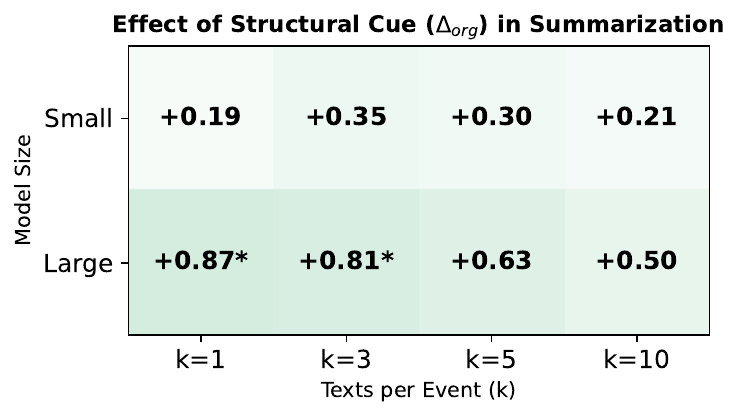}
        \label{fig:subfig3}
        \vspace{-0.5cm}
        \caption{Summarization.}
    \end{subfigure}
    \vspace{-0.4cm}
    \caption{$\Delta_{\text{org}}$ across model scales and document sizes per event ($k$) for each task. * indicates statistical significance ($p < 0.05$).}
    \label{fig:cue_effects_delta}
\end{figure*}

\subsection{Experimental setup}
\subsubsection{Models}
We evaluate instruction-tuned LLMs with varying model sizes and context window lengths. Small models (1--4B) include \texttt{Llama-3.2-1B}, \texttt{Llama-3.2-3B}, \texttt{Gemma-2-2B}, and \texttt{Gemma-3-4B}. Large models (70B+) include \texttt{Llama-3.1-70B}, \texttt{Qwen2.5-72B}, and \texttt{Mistral-Large}~\citep{dubey2024llama3, team2024gemma2, team2025gemma3, yang2024qwen25, jiang2024mistral}.
\footnote{We also test Medium models (7-9B), include \texttt{Llama-3.1-8B}, \texttt{Qwen2.5-7B}, and \texttt{Gemma-2-9B} (See \Cref{sec:additional-results})}
\Cref{tab:model_specs_appendix} summarizes each model's knowledge cutoff. Most models have cutoffs in late 2023 to early 2024, so much of the 2025 data falls beyond these cutoffs, making it unlikely that the models have prior knowledge.

\subsubsection{Implementation Details}
All experiments were conducted using vLLM v0.10.2~\citep{10.1145/3600006.3613165} with temperature 0.0 and random seed 42. We used NVIDIA A100 80GB, B200 180GB and A6000 48GB GPUs. When inputs exceed the model's maximum context length, we uniformly truncate while maintaining proportions across events.

\subsection{Does Structural Cue Help?: Analyzing $\Delta_{org}$}
\Cref{fig:bottleneck} summarizes the effect of structural cues across three tasks. Each bar represents the base performance, effect of structural cues (+ $\Delta_{org}$, green), (- $\Delta_{org}$, red), and the remaining gap to the ceiling ($\Delta_{gap}$, hatched). 
Also, \Cref{fig:cue_effects_delta} show $\Delta_{org}$, where * indicates statistical significance via Wilcoxon signed-rank test ($p < 0.05$); full p-values are reported in \Cref{tab:pvalues}. We report task-specific results, then analyze the difficulties that remain ($\Delta_{gap}$) even with structural cues. 
Full results broken down by model, year, and documents per event ($k$) are provided in \Cref{tab-appendix:topic-clustering,tab-appendix:temporal-qa,tab-appendix:summarization}.

\subsubsection{Topic Clustering: Organization Bottleneck Emerges with Document Quantity}
\label{subsec:clustering}

For small models (1--4B), base $\text{B}^3\text{-F1}$ is 86.03 in $k{=}1$ but drops to 70.04 ($k{=}3$), 68.03 ($k{=}5$), and 67.96 ($k{=}10$). In our setup, clustering is incremental--each arriving document is either assigned to an existing topic or used to create a new one. As $k$ increases, more documents from different topics are mixed together in the context, making correct assignment more challenging.

The structural cue effect follows this pattern. At $k{=}1$, $\Delta_{\text{org}}$ is $-0.25$ (effectively zero), but increases steadily with $k$: $+2.35$ ($k{=}3$), $+3.75$ ($k{=}5$), $+4.37$ ($k{=}10$). When few documents are present, models can distinguish events without cues; as the volume grows, organization becomes a bottleneck. This effect is statistically significant for all $k \geq 3$.

Large models (70B+) show a different pattern. $\Delta_{\text{org}}$ is small and stable regardless of $k$ ($+1.27$ in $k{=}1$, $+0.84$ in $k{=}10$), with base performance in the 82--87 range. Large models maintain their organization capability even under streaming environments.

\subsubsection{Temporal QA: Performance Gains from Explicit Organization}
\label{subsec:qa}

For small models, base accuracy is 72.12 ($k{=}1$), 76.04 ($k{=}3$), 77.19 ($k{=}5$), and 75.39 ($k{=}10$). Unlike clustering, increasing $k$ does not degrade performance, because more documents per event also provide more support for the correct answer.

Despite this, $\Delta_{\text{org}}$ is large and consistent across all conditions: $+9.63$ ($k{=}1$), $+6.01$ ($k{=}3$), $+5.35$ ($k{=}5$), $+6.55$ ($k{=}10$), all statistically significant. Even though more documents per event increase the chance of including answer-relevant information, locating the right information within a context where multiple topics are mixed remains difficult. This is analogous to a needle-in-a-haystack problem: the difficulty is not in the quantity of information but in finding the relevant pieces among heterogeneous content.

This effect is especially large for small models. Large models also show significant $\Delta_{\text{org}}$ ($+3.14$ to $+7.51$), but smaller in magnitude because they already find relevant information reasonably well from raw input. As in \Cref{subsec:clustering}, large models maintain their own organization capability. With cues, large model accuracy reaches the 94--97 range consistently.

\begin{table}[t]
\centering
\caption{Statistical significance of structural cue effect (Wilcoxon signed-rank test). * indicates $p < 0.05$.}
\label{tab:pvalues}
\renewcommand{\arraystretch}{1.2}
\resizebox{\columnwidth}{!}{%
\begin{tabular}{ll|cccc}
\toprule
Task & Scale & $k$=1 & $k$=3 & $k$=5 & $k$=10 \\
\midrule
\multirow{2}{*}{Topic Clustering} & Small & .988 & $<$.001* & .003* & .042* \\
 & Large & $<$.001* & $<$.001* & .001* & .016* \\
\midrule
\multirow{2}{*}{Temporal QA} & Small & $<$.001* & $<$.001* & $<$.001* & $<$.001* \\
 & Large & $<$.001* & $<$.001* & $<$.001* & $<$.001* \\
\midrule
\multirow{2}{*}{Summarization} & Small & .747 & .456 & .345 & .623 \\
 & Large & .008* & .018* & .090 & .265 \\
\bottomrule
\end{tabular}}
\end{table}

\subsubsection{Summarization: Limited Impact of Structural Cues}
\label{subsec:sum}

In contrast to the previous two tasks, $\Delta_{\text{org}}$ for summarization is smaller, remaining below one in all cases. For small models, it ranges from $+0.19$ to $+0.35$; for large models, $+0.50$ to $+0.87$. \Cref{tab:summ_metrics} reports performance across supplementary metrics. While METEOR and ROUGE-2 show relatively larger improvements for large models ($+3.4$ and $+2.3$, respectively), ROUGE-1 and ROUGE-L remain nearly unchanged. Even the largest gains are far smaller than those observed in topic clustering or temporal QA.

\begin{table}[t]
\centering
\caption{Summarization performance across multiple metrics. w/ cue indicates whether structural cues are provided.}
\label{tab:summ_metrics}
\setlength{\tabcolsep}{11pt}
\resizebox{\columnwidth}{!}{%
\begin{tabular}{ll|ccc}
\toprule
Model & Metric & w/o cue & w/ cue & $\Delta_{\text{org}}$ \\
\midrule
\multirow{5}{*}{Small} 
& ROUGE-1 & 26.9 & 26.7 & \phantom{+}0.2 \\
& ROUGE-2 & \phantom{0}8.0 & \phantom{0}9.0 & \phantom{+}1.0 \\
& ROUGE-L & 15.6 & 15.9 & \phantom{+}0.3 \\
& BLEU    & \phantom{0}3.6 & \phantom{0}4.2 & \phantom{+}0.7 \\
& METEOR & 25.3 & 26.6 & \phantom{+}1.4 \\
\midrule
\multirow{5}{*}{Large}
& ROUGE-1 & 26.9 & 27.0 & \phantom{+}0.1 \\
& ROUGE-2 & \phantom{0}8.8 & 11.0 & \phantom{+}2.3 \\
& ROUGE-L & 15.4 & 16.1 & \phantom{+}0.7 \\
& BLEU    & \phantom{0}3.8 & \phantom{0}5.2 & \phantom{+}1.4 \\
& METEOR & 29.8 & 33.2 & \phantom{+}3.4 \\
\bottomrule
\end{tabular}}
\end{table}

\subsection{What Remains Difficult?: Analyzing $\Delta_{gap}$}

The results so far show that structural cues improve performance under certain conditions. Yet even with cues, the gap to the ceiling ($\Delta_{\text{gap}}$) remains substantial. We now analyze what this gap corresponds to in each task.

\subsubsection{Topic Clustering}

We classify clustering errors into over-clustering (more clusters than ground truth), under-clustering (fewer clusters), and exact match.

For small models, cues reduce over-clustering from 34.2\% to 24.3\% ($-9.9$ pp). Cues provide separation signals between events, reducing the error of merging documents from different events into one cluster. The effect is stronger for medium models (16.9\% $\rightarrow$ 0.6\%, $-16.3$ pp).

This improvement comes with a trade-off. The drop in over-clustering is offset by a rise in under-clustering (Small: $+5.6$ pp; Medium: $+16.4$ pp): cues push toward event separation so strongly that documents from the same event are sometimes split into different clusters. Exact match rate improves only modestly for small models (34.6\% $\rightarrow$ 39.0\%) and reaches at most 51.2\% for large models.

For instance, within the same US Presidential Election story, ``Republican debates'' and ``Iowa caucus'' are merged into one cluster, or documents from the same event are split into two clusters based on temporal gaps. Structural cues organize entity-level information by event, but the precise boundary between events requires the model to synthesize cue and context information through its own reasoning--a difficulty that cues alone do not resolve.

\subsubsection{Temporal QA}

As defined in \Cref{subsec:annotation}, QA questions fall into two types: Result Recognition (questions about temporal relationships such as outcomes or causal connections between events) and Entity Tracking (questions about entity state changes over time). Entity Tracking also divides into (1) \texttt{counting}, (2) \texttt{temporal\_order}, (3) \texttt{current\_state}, and (4) \texttt{temporal\_recency}. The cue effect varies clearly by question type.

\paragraph{Types with large cue effects.} (1) \texttt{counting} shows the largest improvement ($+11.1\%$). Questions like ``How many ceasefire violations occurred?'' require aggregating information spread across multiple events; cues make this easier by organizing information per event. (2) \texttt{temporal\_order} also benefits ($+7.8\%$). For questions like ``Did the peace talks happen before or after the election?'', the model must first locate each event in the context before judging their temporal relationship, and cues help with that localization.

\paragraph{Types with limited or no cue effects.} (3) \texttt{current\_state} retains a 21\% error rate even with cues. For example, given ``Who is the current Israeli deputy defence minister?'', the cue provides \texttt{People} $=$ [Avi Dichter, Tzachi Hanegbi, Eli Ben-Dahan, Yoav Galant] for the relevant events, but the model selects Avi Dichter (from an earlier event) over the correct answer, Eli Ben-Dahan. Cues organize who appears in each event, but deciding who \textit{currently} holds the position requires comparing temporal information across events—reasoning that the model must do on its own. This limitation holds across model scales (\texttt{Mistral-Large 123B}: 4.3\% error). (4) \texttt{temporal\_recency} shows a degradation of $-3.7\%$. For questions like ``What is the most recent update?'', the model must judge which information is newest based on dates in the context. When cues make event-level information clearer, the number of candidates grows, making this judgment harder rather than easier. The pattern across types is consistent; questions that require finding and combining information across events benefit from cues, while questions that require temporal reasoning over the found information do not.

\subsubsection{Summarization}
\label{subsubsec:sum-gap-analysis}

Summarization has the largest $\Delta_{\text{gap}}$ of the three tasks ($\sim$84 on the ROUGE-L scale). To identify what remains difficult, we compare summary pairs with and without structural cues from the same condition. Small models with cues include more facts from the source, but struggle to integrate them. For example, a \texttt{Llama-3.2-1B} copies raw cue content (e.g., entity lists) directly into the output, while a \texttt{Llama-3.2-3B} covers more events but lists them without connecting them into a narrative. Cues help models identify what to include, but compressing that information into coherent prose remains an unsolved challenge. Unlike Topic clustering and Temporal QA, summarization is an open-ended generative task whose quality spans multiple dimensions; lexical-based metrics may not fully capture these effects. We further explore this with LLM-as-a-Judge in \Cref{subsec:checkeval}.

\begin{table}[t]
\centering
\small
\caption{Ablation on structural cue components. Each row under w/ cue removes one element to measure its contribution. Bold indicates the best score per setting.}
\label{tab:ablation_cue}
\resizebox{\columnwidth}{!}{%
\begin{tabular}{lccc}
\toprule
\textbf{Setting} & \textbf{Clustering} & \textbf{Temporal QA} & \textbf{Summarization} \\
                 & {\footnotesize (B³-F1)} & {\footnotesize (Accuracy)} & {\footnotesize (ROUGE-L / METEOR)} \\
\midrule
\multicolumn{4}{l}{\textit{\textbf{Gemma-2-2B}}} \\
\midrule
w/o Cue & 71.5 & 80.3 & 15.9 / 26.3 \\
w/ Cue & \textbf{78.0} & \textbf{88.3} & \textbf{16.3} / \textbf{27.7} \\
\midrule
\quad – Location & 76.4 & 87.4 & 16.1 / 26.3 \\
\quad – Event Attrs & 73.6 & 88.1 & \textbf{16.3} / 27.6 \\
\quad – People & 77.8 & 86.9 & 15.9 / 27.1 \\
\quad – Result & 75.6 & 86.3 & 16.2 / 27.4 \\
\midrule
\multicolumn{4}{l}{\textit{\textbf{Qwen2.5-72B}}} \\
\midrule
w/o cue & 82.4 & 89.7 & 14.7 / 29.5 \\
w/ cue & 84.1 & 95.3 & \textbf{15.7} / \textbf{33.1} \\
\midrule
\quad – Location & \textbf{84.5} & \textbf{95.5}& 15.1 / 31.5 \\
\quad – Event Attrs & 83.5 & 96.1 & 15.1 / 31.2 \\
\quad – People & 84.1 & 93.8 & 15.0 / 31.4 \\
\quad – Result & 83.8 & 94.8 & 14.9 / 31.1 \\
\bottomrule
\end{tabular}}
\end{table}

\begin{table}[ht]
\centering
\small
\caption{Effect of cue structure on temporal QA (Acc \%). \textbf{Bold} indicates the best score.}
\label{tab:qa_baseline}
\resizebox{\columnwidth}{!}{%
\begin{tabular}{lcccc}
\toprule
\textbf{Model} & \textbf{w/o cue} & \textbf{RAG} & \textbf{Serialized Facts} & \textbf{w/ cue} \\
\midrule
Gemma-2-2B   & 80.3 & 82.4 & 85.3 & \textbf{88.3} \\
Qwen2.5-72B  & 89.7 & 90.2 & 92.4 & \textbf{95.3} \\
\bottomrule
\end{tabular}}
\end{table}
\section{Further Analysis}

\subsection{Effect of Cue Components}
\label{subsec:cue-analyze}
To analyze which cue elements drive the gains, we conducted an ablation study. The analysis focused on two models showing the most consistent performance gains with structural cues: \texttt{Gemma-2-2B} (Small) and \texttt{Qwen2.5-72B} (Large). We measured performance changes by removing each element in a structural cue (People, Location, Result, EventAttr) one at a time. \Cref{tab:ablation_cue} shows that removing any single component generally degrades performance, confirming that each element contributes to the overall gain. For the small model, Event Attrs and Result show the largest contributions to Topic clustering and Temporal QA, respectively. The large model shows smaller but consistent drops across components. These results indicate that the cue design is reasonably balanced, with each component contributing to at least one task.

\subsection{Effect of Cue Structure}
\label{subsec:cue-structure-analyze}
We further test whether the gains from structural cues come from event-level organization or simply from access to high-quality facts. To separate these, we compare four input conditions on temporal QA using \texttt{Gemma-2-2B} and \texttt{Qwen2.5-72B} (\Cref{tab:qa_baseline}). The w/o cue setting provides the raw document window without any processing. RAG embeds each article with \texttt{all-MiniLM-L6-v2}, scores it against the question by cosine similarity, and keeps only articles above a 0.5 threshold as plain text. Serialized Facts provides the same oracle-selected key facts used in the cues, but as a flat list without event-level grouping. The w/ cue setting is our full condition. RAG barely improves over the w/o cue baseline, serialized facts help more, and the full cue with event-level structure yields the largest gain. This indicates that the improvement comes mostly from how information is organized rather than from fact access alone, which neither retrieval nor unstructured facts reproduce.

\begin{table}[t]
\centering
\caption{Structural cue effect across evaluation metrics for summarization.}
\label{tab:summ_metric_comparison}
\resizebox{\columnwidth}{!}{%
\begin{tabular}{lcccccc}
\toprule
& \multicolumn{3}{c}{Small (1--4B)} & \multicolumn{3}{c}{Large (70B+)} \\
\cmidrule(lr){2-4} \cmidrule(lr){5-7}
Metric & w/o cue & w/ cue & $\Delta_{\text{org}}$ & w/o cue & w/ cue & $\Delta_{\text{org}}$ \\
\midrule
ROUGE-L & 15.6 & 15.9 & +0.3 & 15.4 & 16.1 & +0.7 \\
METEOR & 25.7 & 27.2 & +1.5 & 29.5 & 31.8 & +2.3 \\
CheckEval & 72.4 & 75.4 & +3.0 & 84.8 & 89.7 & +4.9 \\
\bottomrule
\end{tabular}}
\end{table}

\begin{figure}[t]
    \centering
    \includegraphics[width=\columnwidth]{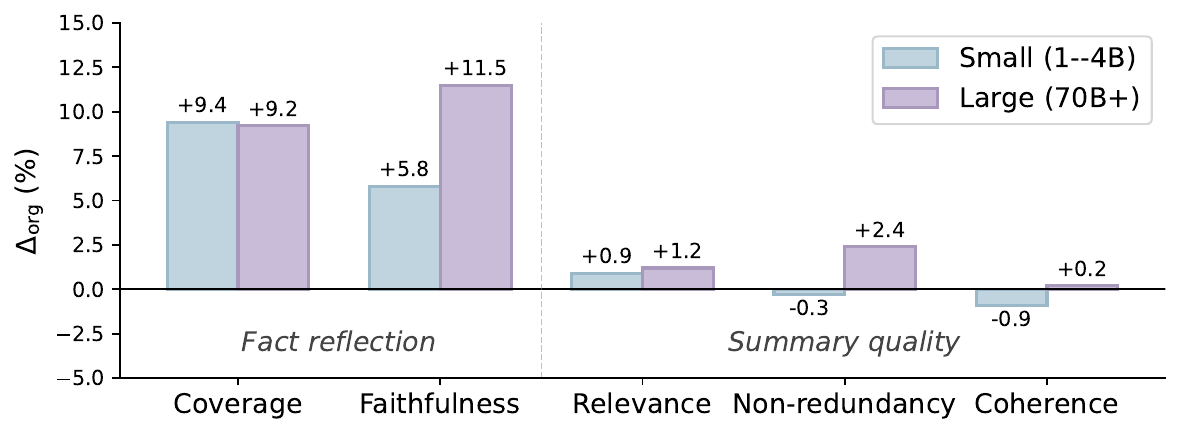}
    \caption{CheckEval results for summarization, with $\Delta_{\text{org}}$ broken down by evaluation dimension.}
    \label{fig:checkeval_breakdown}
\end{figure}

\subsection{Fine-Grained Summarization Analysis with LLM-as-Judge}
\label{subsec:checkeval}

\Cref{subsubsec:sum-gap-analysis} noted that lexical-based metrics may not fully capture effects on individual dimensions. To test this systematically, we use CheckEval~\citep{lee-etal-2025-checkeval}
, which evaluates each quality dimension through decomposed yes-or-no questions and has shown reliable results for summarization compared to conventional LLM-as-judge methods. Based on the qualitative patterns observed in \Cref{subsubsec:sum-gap-analysis}, we assess five dimensions: factual coverage, faithfulness, relevance, non\_redundancy, and coherence. Due to cost constraints, we sample 1,000 paired instances (both without and with cue) from each of seven models (four small and three large, totaling 7,000 pairs) and use \texttt{GPT-4o-mini} as the judge. 

\Cref{tab:summ_metric_comparison} compares the cue effect as detected by ROUGE-L, METEOR, and CheckEval on the same samples. All three metrics show improvement with structural cues, indicating that cues provide some benefit overall. However, directly comparing $\Delta_{\text{org}}$ values across metrics is not appropriate as each captures different aspects of summary quality; what we observe is the consistent trend of improvement. To understand what ROUGE-L may not fully capture, we examine the CheckEval breakdown by evaluation dimension.

Evaluation dimension-level breakdown (\Cref{fig:checkeval_breakdown}) shows two groups. Coverage and faithfulness improve substantially: cues help models identify what should go into the summary. Coherence, non\_redundancy, and relevance show no meaningful change. Breaking this down by model scale, coverage gains are similar for small ($+9.4\%$) and large ($+9.2\%$) models, cues help identify key information regardless of size. Faithfulness gains, however, are roughly twice as large for large models ($+11.5\%$) as for small models ($+5.8\%$): larger models are better at accurately rendering the information they find. Small models tend toward slight coherence degradation. Cues lead them to include more facts, but they lack the capacity to integrate those facts coherently, creating a trade-off between coverage and quality.

\section{Limitation and Future Work}
Benchmarks that evaluate time-sensitive capabilities inevitably become outdated as LLM knowledge cutoffs advance. We release the entire benchmark construction pipeline as a reproducible framework, enabling researchers to add new stories as they emerge or adapt it to other domains.

Our structural cues were constructed offline from complete event information, but in a real streaming setting, such information is not available in advance. One promising direction is to introduce a dedicated component that incrementally constructs and updates structured representations as new documents arrive, whether through tables, knowledge graphs, or other forms of external memory. Our cues captured entity-level organization, but real-world streams also demand tracking deeper structures such as causal chains across events and evolving relationships that require multi-hop reasoning. Beyond organization, our QA analysis shows that temporal reasoning, particularly tracking entity states and judging recency, remains difficult even when organization is fully provided. LLMs need stronger temporal awareness to determine what is current within a given context. Finally, StreamBench currently evaluates each window and task independently, but streaming is naturally sequential: clustering results carry over as new documents arrive, and earlier task outputs can serve as input to later tasks. Supporting such sequential evaluation is another important direction.

\section{Conclusion}
In this paper, we identified two conflicts specific to streaming document environments: intra-topic and inter-topic conflicts. To simulate these challenges in a controlled setting, we constructed StreamBench, a benchmark comprising 605 events and 15,354 documents across three tasks. Using structural cues as a diagnostic probe, we conducted a detailed analysis of where and why models fail. Our analysis shows that the nature of the bottleneck differs across tasks: in topic clustering, organizational difficulty increases with document volume; in temporal QA, locating relevant information across heterogeneous content is the primary challenge; and in summarization, the difficulty lies in compressing and integrating information rather than organizing it. Across all tasks, structural cues consistently help models find and organize information. While reasoning over temporal dynamics remains an open challenge, the clear benefits of organization show that structural cues are a practical and effective starting point. We hope our analysis offers actionable insights for improving how LLMs handle the conflicts inherent in streaming document environments.

\section*{Acknowledgments}
This work was supported by the Institute of Information \& Communications Technology Planning \& Evaluation (IITP)-ICT Creative Consilience Program (IITP-2026-RS-2020-II201819), Information Technology Research Center (IITP-2026-RS-2024-00436857), Artificial Intelligence Star Fellowship Support Program (IITP-2026-RS-2025-02304828), and the National Research Foundation of Korea (NRF) (RS-2026-25494369) funded by the Korea government (MSIT). We additionally thank Takyoung Kim and Jinu Lee for helpful comments on the paper. 



\balance
\bibliographystyle{ACM-Reference-Format}
\bibliography{sample-base}

@article{10.1145/3373464.3373470,
author = {Gomes, Heitor Murilo and Read, Jesse and Bifet, Albert and Barddal, Jean Paul and Gama, Jo\~{a}o},
title = {Machine learning for streaming data: state of the art, challenges, and opportunities},
year = {2019},
issue_date = {December 2019},
publisher = {Association for Computing Machinery},
address = {New York, NY, USA},
volume = {21},
number = {2},
issn = {1931-0145},
url = {https://doi.org/10.1145/3373464.3373470},
doi = {10.1145/3373464.3373470},
journal = {SIGKDD Explor. Newsl.},
month = nov,
pages = {6–22},
numpages = {17}
}

@inbook{10.5555/772260.772262,
author = {Allan, James},
title = {Introduction to topic detection and tracking},
year = {2002},
isbn = {0792376641},
publisher = {Kluwer Academic Publishers},
address = {USA},
booktitle = {Topic Detection and Tracking: Event-Based Information Organization},
pages = {1–16},
numpages = {16}
}

@inproceedings{10.1145/3269206.3269309,
author = {Hoang, Tuan-Anh and Vo, Khoi Duy and Nejdl, Wolfgang},
title = {W2E: A Worldwide-Event Benchmark Dataset for Topic Detection and Tracking},
year = {2018},
isbn = {9781450360142},
publisher = {Association for Computing Machinery},
address = {New York, NY, USA},
url = {https://doi.org/10.1145/3269206.3269309},
doi = {10.1145/3269206.3269309},
booktitle = {Proceedings of the 27th ACM International Conference on Information and Knowledge Management},
pages = {1847–1850},
numpages = {4},
location = {Torino, Italy},
series = {CIKM '18}
}

@inproceedings{nakshatri-etal-2023-using,
    title = "Using {LLM} for Improving Key Event Discovery: Temporal-Guided News Stream Clustering with Event Summaries",
    author = "Nakshatri, Nishanth  and
      Liu, Siyi  and
      Chen, Sihao  and
      Roth, Dan  and
      Goldwasser, Dan  and
      Hopkins, Daniel",
    editor = "Bouamor, Houda  and
      Pino, Juan  and
      Bali, Kalika",
    booktitle = "Findings of the Association for Computational Linguistics: EMNLP 2023",
    month = dec,
    year = "2023",
    address = "Singapore",
    publisher = "Association for Computational Linguistics",
    url = "https://aclanthology.org/2023.findings-emnlp.274/",
    doi = "10.18653/v1/2023.findings-emnlp.274",
    pages = "4162--4173"
}

@inproceedings{zhang-choi-2021-situatedqa,
    title = "{S}ituated{QA}: Incorporating Extra-Linguistic Contexts into {QA}",
    author = "Zhang, Michael  and
      Choi, Eunsol",
    editor = "Moens, Marie-Francine  and
      Huang, Xuanjing  and
      Specia, Lucia  and
      Yih, Scott Wen-tau",
    booktitle = "Proceedings of the 2021 Conference on Empirical Methods in Natural Language Processing",
    month = nov,
    year = "2021",
    address = "Online and Punta Cana, Dominican Republic",
    publisher = "Association for Computational Linguistics",
    url = "https://aclanthology.org/2021.emnlp-main.586/",
    doi = "10.18653/v1/2021.emnlp-main.586",
    pages = "7371--7387"
}

@InProceedings{streamingqa2022,
  title = 	 {{S}treaming{QA}: A Benchmark for Adaptation to New Knowledge over Time in Question Answering Models},
  author =       {Liska, Adam and Kocisky, Tomas and Gribovskaya, Elena and Terzi, Tayfun and Sezener, Eren and Agrawal, Devang and De Masson D'Autume, Cyprien and Scholtes, Tim and Zaheer, Manzil and Young, Susannah and Gilsenan-Mcmahon, Ellen and Austin, Sophia and Blunsom, Phil and Lazaridou, Angeliki},
  booktitle = 	 {Proceedings of the 39th International Conference on Machine Learning},
  pages = 	 {13604--13622},
  year = 	 {2022},
  editor = 	 {Chaudhuri, Kamalika and Jegelka, Stefanie and Song, Le and Szepesvari, Csaba and Niu, Gang and Sabato, Sivan},
  volume = 	 {162},
  series = 	 {Proceedings of Machine Learning Research},
  month = 	 {17--23 Jul},
  publisher =    {PMLR},
  url = 	 {https://proceedings.mlr.press/v162/liska22a.html}
}

@inproceedings{
kasai2023realtime,
title={RealTime {QA}: What's the Answer Right Now?},
author={Jungo Kasai and Keisuke Sakaguchi and yoichi takahashi and Ronan Le Bras and Akari Asai and Xinyan Velocity Yu and Dragomir Radev and Noah A. Smith and Yejin Choi and Kentaro Inui},
booktitle={Thirty-seventh Conference on Neural Information Processing Systems Datasets and Benchmarks Track},
year={2023},
url={https://openreview.net/forum?id=HfKOIPCvsv}
}

@inproceedings{vu-etal-2024-freshllms,
    title = "{F}resh{LLM}s: Refreshing Large Language Models with Search Engine Augmentation",
    author = "Vu, Tu  and
      Iyyer, Mohit  and
      Wang, Xuezhi  and
      Constant, Noah  and
      Wei, Jerry  and
      Wei, Jason  and
      Tar, Chris  and
      Sung, Yun-Hsuan  and
      Zhou, Denny  and
      Le, Quoc  and
      Luong, Thang",
    editor = "Ku, Lun-Wei  and
      Martins, Andre  and
      Srikumar, Vivek",
    booktitle = "Findings of the Association for Computational Linguistics: ACL 2024",
    month = aug,
    year = "2024",
    address = "Bangkok, Thailand",
    publisher = "Association for Computational Linguistics",
    url = "https://aclanthology.org/2024.findings-acl.813/",
    doi = "10.18653/v1/2024.findings-acl.813",
    pages = "13697--13720"
}

@inproceedings{
dai2025are,
title={Are {LLM}s Prescient? A Continuous Evaluation using Daily News as the Oracle},
author={Hui Dai and Ryan Teehan and Mengye Ren},
booktitle={Forty-second International Conference on Machine Learning},
year={2025},
url={https://openreview.net/forum?id=v2nV83Q849}
}

@inproceedings{ouyang-etal-2025-hoh,
    title = "{H}o{H}: A Dynamic Benchmark for Evaluating the Impact of Outdated Information on Retrieval-Augmented Generation",
    author = "Ouyang, Jie  and
      Pan, Tingyue  and
      Cheng, Mingyue  and
      Yan, Ruiran  and
      Luo, Yucong  and
      Lin, Jiaying  and
      Liu, Qi",
    editor = "Che, Wanxiang  and
      Nabende, Joyce  and
      Shutova, Ekaterina  and
      Pilehvar, Mohammad Taher",
    booktitle = "Proceedings of the 63rd Annual Meeting of the Association for Computational Linguistics (Volume 1: Long Papers)",
    month = jul,
    year = "2025",
    address = "Vienna, Austria",
    publisher = "Association for Computational Linguistics",
    url = "https://aclanthology.org/2025.acl-long.301/",
    doi = "10.18653/v1/2025.acl-long.301",
    pages = "6036--6063",
    ISBN = "979-8-89176-251-0"
}

@inproceedings{10.1145/3543507.3583371,
author = {Yoon, Susik and Chan, Hou Pong and Han, Jiawei},
title = {PDSum: Prototype-driven Continuous Summarization of Evolving Multi-document Sets Stream},
year = {2023},
isbn = {9781450394161},
publisher = {Association for Computing Machinery},
address = {New York, NY, USA},
url = {https://doi.org/10.1145/3543507.3583371},
doi = {10.1145/3543507.3583371},
booktitle = {Proceedings of the ACM Web Conference 2023},
pages = {1650–1661},
numpages = {12},
location = {Austin, TX, USA},
series = {WWW '23}
}

@inproceedings{10.1145/3543507.3583507,
author = {Yoon, Susik and Meng, Yu and Lee, Dongha and Han, Jiawei},
title = {SCStory: Self-supervised and Continual Online Story Discovery},
year = {2023},
isbn = {9781450394161},
publisher = {Association for Computing Machinery},
address = {New York, NY, USA},
url = {https://doi.org/10.1145/3543507.3583507},
doi = {10.1145/3543507.3583507},
booktitle = {Proceedings of the ACM Web Conference 2023},
pages = {1853–1864},
numpages = {12},
location = {Austin, TX, USA},
series = {WWW '23}
}

@inproceedings{10.1145/1008992.1009065,
author = {Chieu, Hai Leong and Lee, Yoong Keok},
title = {Query based event extraction along a timeline},
year = {2004},
isbn = {1581138814},
publisher = {Association for Computing Machinery},
address = {New York, NY, USA},
url = {https://doi.org/10.1145/1008992.1009065},
doi = {10.1145/1008992.1009065},
booktitle = {Proceedings of the 27th Annual International ACM SIGIR Conference on Research and Development in Information Retrieval},
pages = {425–432},
numpages = {8},
location = {Sheffield, United Kingdom},
series = {SIGIR '04}
}

@inproceedings{xu-etal-2021-document,
    title = "Document-level Event Extraction via Heterogeneous Graph-based Interaction Model with a Tracker",
    author = "Xu, Runxin  and
      Liu, Tianyu  and
      Li, Lei  and
      Chang, Baobao",
    editor = "Zong, Chengqing  and
      Xia, Fei  and
      Li, Wenjie  and
      Navigli, Roberto",
    booktitle = "Proceedings of the 59th Annual Meeting of the Association for Computational Linguistics and the 11th International Joint Conference on Natural Language Processing (Volume 1: Long Papers)",
    month = aug,
    year = "2021",
    address = "Online",
    publisher = "Association for Computational Linguistics",
    url = "https://aclanthology.org/2021.acl-long.274/",
    doi = "10.18653/v1/2021.acl-long.274",
    pages = "3533--3546"
}

@inproceedings{hu-etal-2024-moments,
    title = "From Moments to Milestones: Incremental Timeline Summarization Leveraging Large Language Models",
    author = "Hu, Qisheng  and
      Moon, Geonsik  and
      Ng, Hwee Tou",
    editor = "Ku, Lun-Wei  and
      Martins, Andre  and
      Srikumar, Vivek",
    booktitle = "Proceedings of the 62nd Annual Meeting of the Association for Computational Linguistics (Volume 1: Long Papers)",
    month = aug,
    year = "2024",
    address = "Bangkok, Thailand",
    publisher = "Association for Computational Linguistics",
    url = "https://aclanthology.org/2024.acl-long.390/",
    doi = "10.18653/v1/2024.acl-long.390",
    pages = "7232--7246"
}

@inproceedings{song-etal-2025-temporal,
    title = "Temporal reasoning for timeline summarisation in social media",
    author = "Song, Jiayu  and
      Akhter, Mahmud Elahi  and
      Atzil-Slonim, Dana  and
      Liakata, Maria",
    editor = "Che, Wanxiang  and
      Nabende, Joyce  and
      Shutova, Ekaterina  and
      Pilehvar, Mohammad Taher",
    booktitle = "Proceedings of the 63rd Annual Meeting of the Association for Computational Linguistics (Volume 1: Long Papers)",
    month = jul,
    year = "2025",
    address = "Vienna, Austria",
    publisher = "Association for Computational Linguistics",
    url = "https://aclanthology.org/2025.acl-long.1362/",
    doi = "10.18653/v1/2025.acl-long.1362",
    pages = "28085--28101",
    ISBN = "979-8-89176-251-0"
}

@inproceedings{zhang-etal-2024-analyzing,
    title = "Analyzing Temporal Complex Events with Large Language Models? A Benchmark towards Temporal, Long Context Understanding",
    author = "Zhang, Zhihan  and
      Cao, Yixin  and
      Ye, Chenchen  and
      Ma, Yunshan  and
      Liao, Lizi  and
      Chua, Tat-Seng",
    editor = "Ku, Lun-Wei  and
      Martins, Andre  and
      Srikumar, Vivek",
    booktitle = "Proceedings of the 62nd Annual Meeting of the Association for Computational Linguistics (Volume 1: Long Papers)",
    month = aug,
    year = "2024",
    address = "Bangkok, Thailand",
    publisher = "Association for Computational Linguistics",
    url = "https://aclanthology.org/2024.acl-long.87/",
    doi = "10.18653/v1/2024.acl-long.87",
    pages = "1588--1606"
}

@inproceedings{du-etal-2025-context,
    title = "Context Length Alone Hurts {LLM} Performance Despite Perfect Retrieval",
    author = "Du, Yufeng  and
      Tian, Minyang  and
      Ronanki, Srikanth  and
      Rongali, Subendhu  and
      Bodapati, Sravan Babu  and
      Galstyan, Aram  and
      Wells, Azton  and
      Schwartz, Roy  and
      Huerta, Eliu A  and
      Peng, Hao",
    editor = "Christodoulopoulos, Christos  and
      Chakraborty, Tanmoy  and
      Rose, Carolyn  and
      Peng, Violet",
    booktitle = "Findings of the Association for Computational Linguistics: EMNLP 2025",
    month = nov,
    year = "2025",
    address = "Suzhou, China",
    publisher = "Association for Computational Linguistics",
    url = "https://aclanthology.org/2025.findings-emnlp.1264/",
    doi = "10.18653/v1/2025.findings-emnlp.1264",
    pages = "23281--23298",
    ISBN = "979-8-89176-335-7",
}

@inproceedings{lee-etal-2025-checkeval,
    title = "{C}heck{E}val: A reliable {LLM}-as-a-Judge framework for evaluating text generation using checklists",
    author = "Lee, Yukyung  and
      Kim, JoongHoon  and
      Kim, Jaehee  and
      Cho, Hyowon  and
      Kang, Jaewook  and
      Kang, Pilsung  and
      Kim, Najoung",
    editor = "Christodoulopoulos, Christos  and
      Chakraborty, Tanmoy  and
      Rose, Carolyn  and
      Peng, Violet",
    booktitle = "Proceedings of the 2025 Conference on Empirical Methods in Natural Language Processing",
    month = nov,
    year = "2025",
    address = "Suzhou, China",
    publisher = "Association for Computational Linguistics",
    url = "https://aclanthology.org/2025.emnlp-main.796/",
    doi = "10.18653/v1/2025.emnlp-main.796",
    pages = "15771--15798",
    ISBN = "979-8-89176-332-6"
}

@article{dubey2024llama3,
  title={The LLaMA 3 Herd of Models},
  author={Abhimanyu Dubey and Abhinav Jauhri and Abhinav Pandey and Abhishek Kadian and Ahmad Al-Dahle and Aiesha Letman and Akhil Mathur and Alan Schelten and Amy Yang and Angela Fan and others},
  journal={arxiv},
  year={2024},
  volume={abs/2407.21783},
  url={https://arxiv.org/abs/2407.21783}
}

@article{team2024gemma2,
  title={Gemma 2: Improving open language models at a practical size},
  author={Gemma Team and Morgane Riviere and Shreya Pathak and Pier Giuseppe Sessa and Cassidy Hardin and Surya Bhupatiraju and L{\'e}onard Hussenot and Thomas Mesnard and Bobak Shahriari and Alexandre Ram{\'e} and others},
  journal={arxiv},
  year={2024},
  volume={abs/2408.00118},
  url={https://arxiv.org/abs/2408.00118}
}

@article{team2025gemma3,
  title={Gemma 3 technical report},
  author={Team, Gemma and Kamath, Aishwarya and Ferret, Johan and Pathak, Shreya and Vieillard, Nino and Merhej, Ramona and Perrin, Sarah and Matejovicova, Tatiana and Ram{\'e}, Alexandre and Rivi{\`e}re, Morgane and others},
  journal={arxiv},
  year={2025},
  volume={abs/2503.19786},
  url={https://arxiv.org/abs/2503.19786}
}

@article{yang2024qwen25,
  title={Qwen2.5 Technical Report},
  author={Yang, An and Yang, Baosong and Zhang, Beichen and Hui, Binyuan and Zheng, Bo and Yu, Bowen and Li, Chengyuan and Liu, Dayiheng and Huang, Fei and Wei, Haoran and others},
  journal={arxiv},
  year={2024},
  volume={abs/2412.15115},
  url={https://arxiv.org/abs/2412.15115}
}

@article{jiang2024mistral,
  title={Mistral 7B. arXiv 2023},
  author={Jiang, AQ and Sablayrolles, A and Mensch, A and Bamford, C and Chaplot, DS and Casas, Ddl and Bressand, F and Lengyel, G and Lample, G and Saulnier, L and others},
  journal={arxiv},
  year={2024},
  volume={abs/2310.06825},
  url={https://arxiv.org/abs/2310.06825}
}

@article{achiam2023gpt4,
  title={Gpt-4 technical report},
  author={Achiam, Josh and Adler, Steven and Agarwal, Sandhini and Ahmad, Lama and Akkaya, Ilge and Aleman, Florencia Leoni and Almeida, Diogo and Altenschmidt, Janko and Altman, Sam and Anadkat, Shyamal and others},
  journal={arxiv},
  year={2023},
  volume={abs/2303.08774},
  url={https://arxiv.org/abs/2303.08774}
}

@inproceedings{10.1145/3600006.3613165,
author = {Kwon, Woosuk and Li, Zhuohan and Zhuang, Siyuan and Sheng, Ying and Zheng, Lianmin and Yu, Cody Hao and Gonzalez, Joseph and Zhang, Hao and Stoica, Ion},
title = {Efficient Memory Management for Large Language Model Serving with PagedAttention},
year = {2023},
isbn = {9798400702297},
publisher = {Association for Computing Machinery},
address = {New York, NY, USA},
url = {https://doi.org/10.1145/3600006.3613165},
doi = {10.1145/3600006.3613165},
booktitle = {Proceedings of the 29th Symposium on Operating Systems Principles},
pages = {611–626},
numpages = {16},
location = {Koblenz, Germany},
series = {SOSP '23}
}

@inproceedings{levy-etal-2024-task,
    title = "Same Task, More Tokens: the Impact of Input Length on the Reasoning Performance of Large Language Models",
    author = "Levy, Mosh  and
      Jacoby, Alon  and
      Goldberg, Yoav",
    editor = "Ku, Lun-Wei  and
      Martins, Andre  and
      Srikumar, Vivek",
    booktitle = "Proceedings of the 62nd Annual Meeting of the Association for Computational Linguistics (Volume 1: Long Papers)",
    month = aug,
    year = "2024",
    address = "Bangkok, Thailand",
    publisher = "Association for Computational Linguistics",
    url = "https://aclanthology.org/2024.acl-long.818/",
    doi = "10.18653/v1/2024.acl-long.818",
    pages = "15339--15353"
}

@article{li2025longcontext,
    title={Long-context {LLM}s Struggle with Long In-context Learning},
    author={Tianle Li and Ge Zhang and Quy Duc Do and Xiang Yue and Wenhu Chen},
    journal={Transactions on Machine Learning Research},
    issn={2835-8856},
    year={2025},
    url={https://openreview.net/forum?id=Cw2xlg0e46},
    note={}
}

@article{fabbri-etal-2021-summeval,
    title = "{S}umm{E}val: Re-evaluating Summarization Evaluation",
    author = "Fabbri, Alexander R.  and
      Kry{\'s}ci{\'n}ski, Wojciech  and
      McCann, Bryan  and
      Xiong, Caiming  and
      Socher, Richard  and
      Radev, Dragomir",
    editor = "Roark, Brian  and
      Nenkova, Ani",
    journal = "Transactions of the Association for Computational Linguistics",
    volume = "9",
    year = "2021",
    address = "Cambridge, MA",
    publisher = "MIT Press",
    url = "https://aclanthology.org/2021.tacl-1.24/",
    doi = "10.1162/tacl_a_00373",
    pages = "391--409",
}

@inproceedings{lin2004rouge,
  title={Rouge: A package for automatic evaluation of summaries},
  author={Lin, Chin-Yew},
  booktitle={Text summarization branches out},
  pages={74--81},
  year={2004}
}

@article{zhang2019bertscore,
  title={Bertscore: Evaluating text generation with bert},
  author={Zhang, Tianyi and Kishore, Varsha and Wu, Felix and Weinberger, Kilian Q and Artzi, Yoav},
  journal={arXiv:1904.09675},
  year={2019}
}

@inproceedings{banerjee-lavie-2005-meteor,
    title = "{METEOR}: An Automatic Metric for {MT} Evaluation with Improved Correlation with Human Judgments",
    author = "Banerjee, Satanjeev  and
      Lavie, Alon",
    editor = "Goldstein, Jade  and
      Lavie, Alon  and
      Lin, Chin-Yew  and
      Voss, Clare",
    booktitle = "Proceedings of the {ACL} Workshop on Intrinsic and Extrinsic Evaluation Measures for Machine Translation and/or Summarization",
    month = jun,
    year = "2005",
    address = "Ann Arbor, Michigan",
    publisher = "Association for Computational Linguistics",
    url = "https://aclanthology.org/W05-0909/",
    pages = "65--72"
}

@inproceedings{papineni2002bleu,
  title={Bleu: a method for automatic evaluation of machine translation},
  author={Papineni, Kishore and Roukos, Salim and Ward, Todd and Zhu, Wei-Jing},
  booktitle={Proceedings of the 40th annual meeting of the Association for Computational Linguistics},
  pages={311--318},
  year={2002}
}

@inproceedings{reimers-2019-sentence-bert,
    title = "Sentence-BERT: Sentence Embeddings using Siamese BERT-Networks",
    author = "Reimers, Nils and Gurevych, Iryna",
    booktitle = "Proceedings of the 2019 Conference on Empirical Methods in Natural Language Processing",
    month = "11",
    year = "2019",
    publisher = "Association for Computational Linguistics",
    url = "https://arxiv.org/abs/1908.10084",
}

@article{10.1145/3704922,
author = {Garcia, Cristiano Mesquita and Abilio, Ramon and Koerich, Alessandro Lameiras and Britto, Alceu de Souza and Barddal, Jean Paul},
title = {Concept Drift Adaptation in Text Stream Mining Settings: A Systematic Review},
year = {2025},
issue_date = {April 2025},
publisher = {Association for Computing Machinery},
address = {New York, NY, USA},
volume = {16},
number = {2},
issn = {2157-6904},
url = {https://doi.org/10.1145/3704922},
doi = {10.1145/3704922},
journal = {ACM Trans. Intell. Syst. Technol.},
month = feb,
articleno = {27},
numpages = {67}
}

@article{10.1109/TKDE.2024.3352100,
author = {Pan, Shirui and Luo, Linhao and Wang, Yufei and Chen, Chen and Wang, Jiapu and Wu, Xindong},
title = {Unifying Large Language Models and Knowledge Graphs: A Roadmap},
year = {2024},
issue_date = {July 2024},
publisher = {IEEE Educational Activities Department},
address = {USA},
volume = {36},
number = {7},
issn = {1041-4347},
url = {https://doi.org/10.1109/TKDE.2024.3352100},
doi = {10.1109/TKDE.2024.3352100},
journal = {IEEE Trans. on Knowl. and Data Eng.},
month = jul,
pages = {3580–3599},
numpages = {20}
}

@article{wu2024thinking,
  title={Thinking with knowledge graphs: Enhancing LLM reasoning through structured data},
  author={Wu, Xue and Tsioutsiouliklis, Kostas},
  journal={arXiv preprint arXiv:2412.10654},
  year={2024}
}

\clearpage
\appendix
\begin{figure}[ht]
    \centering
    \includegraphics[width=\columnwidth]{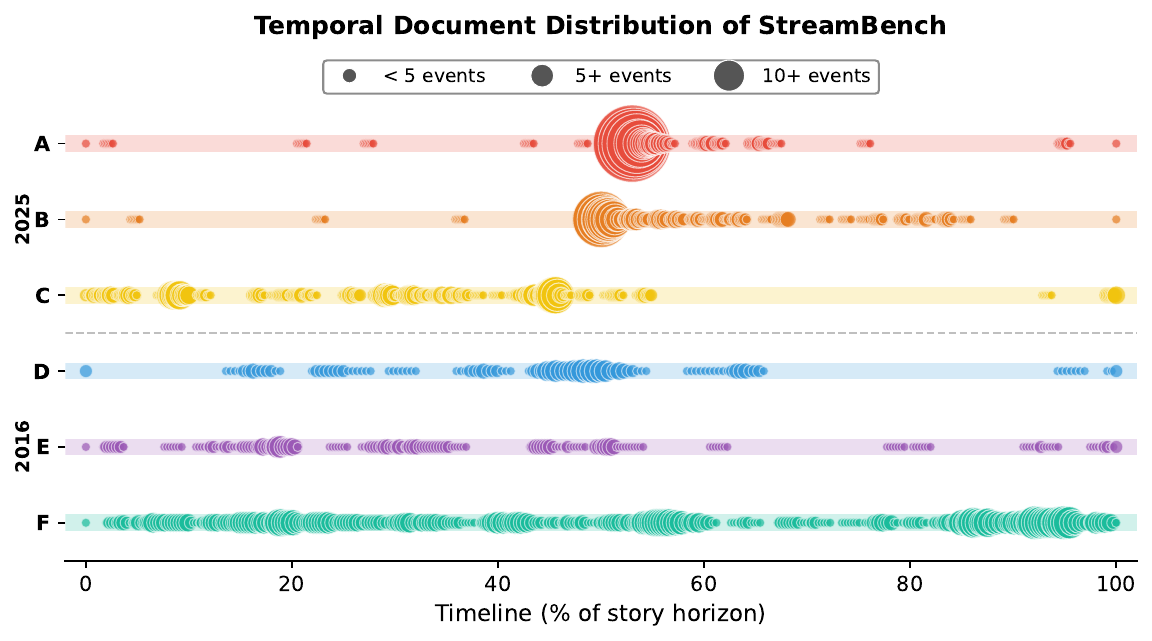}
    \caption{Temporal Document distribution of StreamBench. Each row represents a story (A–F), with bubbles indicating active 7-day windows. Bubble size reflects the number of events per window.}
    \label{fig:time-distribution}
\end{figure}

\begin{table}[ht]
\centering
\small
\caption{Model specifications and knowledge cutoff dates. $^*$ denotes an approximate cutoff, as it is not officially documented by the provider.}
\label{tab:model_specs_appendix}
\resizebox{\columnwidth}{!}{%
\begin{tabular}{llcc}
\toprule
\textbf{Size} & \textbf{Model} & \textbf{Params} & \textbf{Cutoff} \\
\midrule
\multirow{4}{*}{\textbf{Small}}
& \href{https://huggingface.co/meta-Llama/Llama-3.2-1B-Instruct}{Llama-3.2-1B-Instruct} & 1B & Dec 2023 \\
& \href{https://huggingface.co/meta-Llama/Llama-3.2-3B-Instruct}{Llama-3.2-3B-Instruct} & 3B & Dec 2023 \\
& \href{https://huggingface.co/google/gemma-2-2b-it}{Gemma-2-2B-it} & 2B & Jun 2024 \\
& \href{https://huggingface.co/google/gemma-3-4b-it}{Gemma-3-4B-it} & 4B & Aug 2024 \\
\midrule
\multirow{3}{*}{\textbf{Medium}}
& \href{https://huggingface.co/meta-Llama/Llama-3.1-8B-Instruct}{Llama-3.1-8B-Instruct} & 8B & Dec 2023 \\
& \href{https://huggingface.co/Qwen/Qwen2.5-7B-Instruct}{Qwen2.5-7B-Instruct} & 7B & Oct 2023$^*$ \\
& \href{https://huggingface.co/google/gemma-2-9b-it}{Gemma-2-9B-it} & 9B & Jun 2024 \\
\midrule
\multirow{3}{*}{\textbf{Large}}
& \href{https://huggingface.co/meta-Llama/Llama-3.1-70B-Instruct}{Llama-3.1-70B-Instruct} & 70B & Dec 2023 \\
& \href{https://huggingface.co/Qwen/Qwen2.5-72B-Instruct}{Qwen2.5-72B-Instruct} & 72B & Oct 2023$^*$ \\
& \href{https://huggingface.co/mistralai/Mistral-Large-Instruct-2411}{Mistral-Large-Instruct-2411} & 123B & Oct 2023$^*$ \\
\bottomrule
\end{tabular}}
\end{table}

\begin{table*}[ht]
\centering
\caption{Model and year specific topic clustering performance (B$^3$ F1) across temporal cue conditions and documents per event ($k$). \textbf{Bold} indicates the best score in each column.}
\label{tab-appendix:topic-clustering}
\resizebox{0.95\textwidth}{!}{%
\begin{tabular}{cl cccccccccccccccc}
\toprule
\textbf{Size} & \textbf{Model} & \multicolumn{4}{c}{$k$=1} & \multicolumn{4}{c}{$k$=3} & \multicolumn{4}{c}{$k$=5} & \multicolumn{4}{c}{$k$=10} \\
\cmidrule(lr){3-6} \cmidrule(lr){7-10} \cmidrule(lr){11-14} \cmidrule(lr){15-18} 
 &  & \multicolumn{2}{c}{2016} & \multicolumn{2}{c}{2025} & \multicolumn{2}{c}{2016} & \multicolumn{2}{c}{2025} & \multicolumn{2}{c}{2016} & \multicolumn{2}{c}{2025} & \multicolumn{2}{c}{2016} & \multicolumn{2}{c}{2025} \\
\cmidrule(lr){3-4} \cmidrule(lr){5-6} \cmidrule(lr){7-8} \cmidrule(lr){9-10} \cmidrule(lr){11-12} \cmidrule(lr){13-14} \cmidrule(lr){15-16} \cmidrule(lr){17-18} 
 &  & \small wo/ cue & \small w/ cue & \small wo/ cue & \small w/ cue & \small wo/ cue & \small w/ cue & \small wo/ cue & \small w/ cue & \small wo/ cue & \small w/ cue & \small wo/ cue & \small w/ cue & \small wo/ cue & \small w/ cue & \small wo/ cue & \small w/ cue \\
\midrule
\multirow{4}{*}{Small} & Llama-3.2-1B & 88.37 & 86.19 & 88.98 & \textbf{88.93} & 63.65 & 66.85 & 70.88 & 71.25 & 63.45 & 68.38 & 72.78 & 72.79 & 66.43 & 71.28 & 76.05 & 75.60 \\
 & Gemma-2-2B & \textbf{92.92} & 91.10 & \textbf{91.92} & 85.90 & 58.28 & 64.00 & 70.28 & 74.54 & 51.49 & 64.71 & 63.33 & 74.58 & 49.06 & 68.71 & 60.55 & 76.89 \\
 & Llama-3.2-3B & 86.35 & 87.69 & 84.91 & 88.73 & 67.19 & 71.02 & 72.98 & 72.50 & 64.01 & 65.41 & 70.03 & 67.80 & 61.02 & 59.13 & 68.44 & 63.21 \\
 & Gemma-3-4B & 77.37 & 78.10 & 77.40 & 79.54 & 77.99 & 78.63 & 79.08 & 80.33 & 78.87 & 79.11 & 80.25 & 81.40 & 80.19 & 80.53 & 81.91 & 83.25 \\
\midrule
\multirow{3}{*}{Medium} & Qwen2.5-7B & 80.76 & 77.44 & 83.24 & 79.60 & 81.83 & 78.26 & 80.61 & 79.64 & 81.74 & 78.83 & 80.66 & 80.78 & 82.50 & 80.35 & 78.76 & 81.68 \\
 & Llama-3.1-8B & 83.72 & 77.40 & 81.97 & 79.12 & 78.01 & 78.10 & 79.74 & 80.15 & 75.61 & 78.83 & 78.91 & 80.86 & 68.63 & 79.82 & 79.15 & 82.49 \\
 & Gemma-2-9B & 77.49 & 77.90 & 78.41 & 78.64 & 78.25 & 78.40 & 79.50 & 79.92 & 78.86 & 79.30 & 80.42 & 80.83 & 80.47 & 80.70 & 82.23 & 82.29 \\
\midrule
\multirow{3}{*}{Large} & Llama-3.1-70B & 83.68 & 84.71 & 83.54 & 84.43 & 82.43 & 83.62 & 81.39 & 81.53 & 82.36 & 84.09 & 81.90 & 82.53 & \textbf{82.93} & \textbf{84.50} & \textbf{82.43} & 82.74 \\
 & Qwen2.5-72B & 89.92 & 91.28 & 85.50 & 86.96 & \textbf{84.72} & \textbf{84.96} & 81.33 & 82.95 & 82.71 & \textbf{84.76} & 81.53 & \textbf{83.05} & 82.68 & 83.08 & 81.42 & \textbf{83.36} \\
 & Mistral-Large & 90.54 & \textbf{91.90} & 87.03 & 88.53 & 84.34 & 84.91 & \textbf{82.71} & \textbf{83.61} & \textbf{83.18} & 83.22 & \textbf{82.59} & 82.83 & 81.74 & 81.72 & 82.13 & 82.99 \\
\bottomrule
\end{tabular}}
\end{table*}

\begin{table*}[ht]
\centering
\caption{Model and year specific temporal question answering accuracy across temporal cue conditions and documents per event ($k$). \textbf{Bold} indicates the best score in each column.}
\label{tab-appendix:temporal-qa}
\resizebox{0.95\textwidth}{!}{%
\begin{tabular}{cl cccccccccccccccc}
\toprule
\textbf{Size} & \textbf{Model} & \multicolumn{4}{c}{$k$=1} & \multicolumn{4}{c}{$k$=3} & \multicolumn{4}{c}{$k$=5} & \multicolumn{4}{c}{$k$=10} \\
\cmidrule(lr){3-6} \cmidrule(lr){7-10} \cmidrule(lr){11-14} \cmidrule(lr){15-18} 
 &  & \multicolumn{2}{c}{2016} & \multicolumn{2}{c}{2025} & \multicolumn{2}{c}{2016} & \multicolumn{2}{c}{2025} & \multicolumn{2}{c}{2016} & \multicolumn{2}{c}{2025} & \multicolumn{2}{c}{2016} & \multicolumn{2}{c}{2025} \\
\cmidrule(lr){3-4} \cmidrule(lr){5-6} \cmidrule(lr){7-8} \cmidrule(lr){9-10} \cmidrule(lr){11-12} \cmidrule(lr){13-14} \cmidrule(lr){15-16} \cmidrule(lr){17-18} 
 &  & \small wo/ cue & \small w/ cue & \small wo/ cue & \small w/ cue & \small wo/ cue & \small w/ cue & \small wo/ cue & \small w/ cue & \small wo/ cue & \small w/ cue & \small wo/ cue & \small w/ cue & \small wo/ cue & \small w/ cue & \small wo/ cue & \small w/ cue \\
\midrule
\multirow{4}{*}{Small} & Llama-3.2-1B & 42.92 & 50.74 & 53.88 & 66.25 & 44.45 & 52.55 & 55.89 & 64.72 & 49.27 & 53.28 & 55.27 & 62.47 & 45.96 & 53.07 & 53.79 & 60.76 \\
 & Gemma-2-2B & 76.72 & 87.05 & 78.45 & 90.52 & 78.12 & 87.86 & 79.78 & 86.65 & 78.55 & 88.04 & 81.99 & 87.89 & 74.23 & 87.08 & 80.85 & 87.95 \\
 & Llama-3.2-3B & 80.61 & 90.42 & 81.32 & 92.41 & 86.82 & 90.54 & 87.50 & 93.53 & 88.01 & 92.30 & 87.02 & 93.13 & 86.58 & 90.31 & 86.36 & 92.81 \\
 & Gemma-3-4B & 80.95 & 88.47 & 82.10 & 88.11 & 85.27 & 87.42 & 90.51 & 93.19 & 86.22 & 90.28 & 91.17 & 92.95 & 84.72 & 88.15 & 90.64 & 95.39 \\
\midrule
\multirow{3}{*}{Medium} & Qwen2.5-7B & 85.22 & 93.32 & \textbf{89.52} & \textbf{97.77} & 93.40 & 96.67 & 90.23 & 94.40 & 91.26 & 94.10 & 87.92 & 93.37 & 89.49 & 94.15 & 86.86 & 94.30 \\
 & Llama-3.1-8B & 82.15 & 88.94 & 76.12 & 80.25 & 85.37 & 91.06 & 74.83 & 82.15 & 84.12 & 89.73 & 75.72 & 79.84 & 83.28 & 89.41 & 75.45 & 80.15 \\
 & Gemma-2-9B & 23.68 & 23.80 & 29.29 & 31.26 & 24.12 & 24.12 & 28.95 & 30.74 & 24.13 & 24.13 & 28.78 & 30.95 & 24.07 & 24.50 & 28.74 & 29.83 \\
\midrule
\multirow{3}{*}{Large} & Llama-3.1-70B & 87.99 & 93.65 & 84.76 & 93.17 & 91.80 & 93.33 & 89.82 & 93.61 & 92.77 & 95.30 & 87.85 & 93.12 & 92.47 & 93.75 & 87.36 & 90.56 \\
 & Qwen2.5-72B & 88.41 & 96.41 & 86.00 & 94.68 & 93.22 & \textbf{96.96} & \textbf{91.76} & 95.43 & 92.76 & 96.72 & 90.10 & \textbf{95.70} & 92.05 & 97.06 & \textbf{91.04} & \textbf{95.55} \\
 & Mistral-Large & \textbf{92.01} & \textbf{97.09} & 85.17 & 94.41 & \textbf{94.45} & 96.69 & 91.69 & \textbf{95.55} & \textbf{95.02} & \textbf{97.49} & \textbf{91.49} & 94.39 & \textbf{94.39} & \textbf{97.90} & 90.95 & 95.01 \\
\bottomrule
\end{tabular}}
\end{table*}

\section{Temporal QA Generation Protocol}
\label{app:temporal-qa-gen}

\subsection{Question Types}
We generated multiple-choice questions for events with sufficient structured cues. Based on the type of information required, we categorize QA pairs into two classes. Result Recognition questions (e.g., "What was the result of [event]?") require reasoning over causal relationships across temporally distinct events, while Entity Tracking questions (e.g., "Who/Where is currently [role/status]?") require tracking entity states over time and resolving conflicts by prioritizing recent information. Question type distribution consists of 623 Result Recognition questions (57.3\%) and 464 Entity Tracking questions (42.7\%). 

\subsection{Generation Constraints}
QA generation jointly considers questions, answers, and multiple-choice options under strict constraints. Answers are selected from structured cue fields associated with each event (e.g., Result, People, Location) and must be supported by the referenced articles, with specificity aligned to the reported information. During question generation, we avoid including answer strings or lexically identical phrasing from source articles, exclude factual attributes (e.g., birthplace), and ensure that questions are tied to a temporal reference for clarity. Choices are constructed through a two-stage process: we first generate a pool of 10 plausible candidates and then select a balanced subset of distractors. All options maintain consistent specificity and remain contextually plausible within the event window.

\subsection{Verification}
All QA pairs are verified using the same constraint set applied during initial QA generation. We first perform large-scale automated verification using GPT-4o, which checks each QA pair for constraint compliance. Out of 1,483 initially generated QA pairs, 1,087 satisfy all constraints and receive perfect scores, and are retained in the final dataset.
Human verification on 108 random samples (10\%) showed 80.6-83.3\% agreement with automatic validation. 

\section{Additional Results}
\label{sec:additional-results}

\Cref{tab-appendix:topic-clustering} reports topic clustering performance (B$^3$ F1) by model, year, and documents per event ($k$). \Cref{tab-appendix:temporal-qa} reports temporal QA accuracy under the same breakdown. For summarization, \Cref{tab-appendix:summarization} reports ROUGE scores and METEOR scores. In~\Cref{tab-appendix:temporal-qa}, Gemma-2-9B shows notably low QA accuracy (23--31\%) compared to other medium-scale models. Manual inspection confirmed that this is due to formatting failures: the model does not output answers in the required multiple-choice format, making most responses unparseable.

\begin{table*}[t]
\centering
\caption{Model and year specific multi-document summarization performance (ROUGE-L and METEOR) across temporal cue conditions and documents per event ($k$). \textbf{Bold} indicates the best score in each column within each metric block.}
\label{tab-appendix:summarization}
\resizebox{0.95\textwidth}{!}{%
\begin{tabular}{cl cccccccccccccccc}
\toprule
\textbf{Size} & \textbf{Model} & \multicolumn{4}{c}{$k$=1} & \multicolumn{4}{c}{$k$=3} & \multicolumn{4}{c}{$k$=5} & \multicolumn{4}{c}{$k$=10} \\
\cmidrule(lr){3-6} \cmidrule(lr){7-10} \cmidrule(lr){11-14} \cmidrule(lr){15-18} 
 &  & \multicolumn{2}{c}{2016} & \multicolumn{2}{c}{2025} & \multicolumn{2}{c}{2016} & \multicolumn{2}{c}{2025} & \multicolumn{2}{c}{2016} & \multicolumn{2}{c}{2025} & \multicolumn{2}{c}{2016} & \multicolumn{2}{c}{2025} \\
\cmidrule(lr){3-4} \cmidrule(lr){5-6} \cmidrule(lr){7-8} \cmidrule(lr){9-10} \cmidrule(lr){11-12} \cmidrule(lr){13-14} \cmidrule(lr){15-16} \cmidrule(lr){17-18} 
 &  & \small wo/ cue & \small w/ cue & \small wo/ cue & \small w/ cue & \small wo/ cue & \small w/ cue & \small wo/ cue & \small w/ cue & \small wo/ cue & \small w/ cue & \small wo/ cue & \small w/ cue & \small wo/ cue & \small w/ cue & \small wo/ cue & \small w/ cue \\
\midrule
\multicolumn{18}{l}{\textit{ROUGE-L}} \\
\midrule
\multirow{4}{*}{Small} & Llama-3.2-1B & 16.69 & 17.42 & 13.51 & 12.30 & 17.30 & 18.68 & 13.21 & 12.52 & 17.59 & 18.43 & 13.09 & 12.52 & 17.94 & 18.09 & 13.22 & 12.33 \\
 & Gemma-2-2B & 12.81 & 10.31 & 15.80 & 15.31 & 8.63 & 7.84 & 15.89 & 15.33 & 7.42 & 7.31 & 15.97 & 15.15 & 6.90 & 6.79 & 15.90 & 15.38 \\
 & Llama-3.2-3B & 17.92 & 18.97 & 14.96 & 14.92 & 18.37 & 19.35 & 15.26 & 15.19 & 18.32 & 19.18 & 15.55 & 15.39 & 18.43 & 19.32 & 15.52 & 15.61 \\
 & Gemma-3-4B & 17.37 & 20.56 & \textbf{18.15} & \textbf{18.96} & 17.17 & 19.53 & \textbf{18.21} & \textbf{18.39} & 17.13 & 19.38 & \textbf{18.37} & \textbf{18.48} & 17.36 & 19.41 & \textbf{18.91} & \textbf{18.92} \\
\midrule
\multirow{3}{*}{Medium} & Qwen2.5-7B & 18.02 & \textbf{20.83} & 13.13 & 13.68 & \textbf{18.38} & \textbf{20.43} & 13.54 & 13.73 & \textbf{18.47} & \textbf{21.02} & 13.61 & 13.82 & \textbf{18.63} & \textbf{20.95} & 13.77 & 13.86 \\
 & Llama-3.1-8B & 17.60 & 18.66 & 11.15 & 12.65 & 17.91 & 19.32 & 11.89 & 12.83 & 18.20 & 19.03 & 11.94 & 12.95 & 18.43 & 19.93 & 11.75 & 12.80 \\
 & Gemma-2-9B & \textbf{18.04} & 16.73 & 16.13 & 15.26 & 17.32 & 16.04 & 16.78 & 16.57 & 17.21 & 16.09 & 17.28 & 16.51 & 16.93 & 16.33 & 17.50 & 16.83 \\
\midrule
\multirow{3}{*}{Large} & Llama-3.1-70B & 14.51 & 15.01 & 15.33 & 14.57 & 14.49 & 14.33 & 15.31 & 14.71 & 14.51 & 14.21 & 15.55 & 14.48 & 14.85 & 14.07 & 15.64 & 14.56 \\
 & Qwen2.5-72B & 13.34 & 15.32 & 14.71 & 15.44 & 13.11 & 15.36 & 14.58 & 15.66 & 13.25 & 15.32 & 14.69 & 15.88 & 13.30 & 15.54 & 14.80 & 15.93 \\
 & Mistral-Large & 17.53 & 19.40 & 18.01 & 18.91 & 16.62 & 17.98 & 17.21 & 18.13 & 16.50 & 17.69 & 17.16 & 17.84 & 16.69 & 17.75 & 17.42 & 17.86 \\
\midrule
\multicolumn{18}{l}{\textit{METEOR}} \\
\midrule
\multirow{4}{*}{Small} & Llama-3.2-1B & 24.16 & 26.12 & 23.85 & 23.96 & 25.41 & 26.90 & 23.58 & 24.22 & 25.71 & 26.75 & 23.74 & 24.70 & 26.44 & 26.32 & 23.83 & 24.22 \\
 & Gemma-2-2B & 21.91 & 17.52 & 26.13 & 27.58 & 16.16 & 11.13 & 26.58 & 27.83 & 13.49 & 10.39 & 26.35 & 27.62 & 11.90 & 9.37 & 25.98 & 27.64 \\
 & Llama-3.2-3B & 26.26 & 28.67 & 27.51 & 28.77 & 27.51 & 29.35 & 28.73 & 29.71 & 27.53 & 28.80 & 28.63 & 29.56 & 27.69 & 28.74 & 28.59 & 29.92 \\
 & Gemma-3-4B & 25.84 & 31.51 & 28.26 & 33.81 & 26.34 & 30.33 & 28.70 & 33.14 & 25.90 & 30.07 & 28.83 & 33.03 & 26.43 & 30.66 & 28.85 & 33.27 \\
\midrule
\multirow{3}{*}{Medium} & Qwen2.5-7B & 26.44 & 32.01 & 26.18 & 29.61 & 27.83 & 31.90 & 27.30 & 29.68 & 27.96 & 32.01 & 27.66 & 29.82 & 27.97 & 32.07 & 28.01 & 29.63 \\
 & Llama-3.1-8B & 25.98 & 28.04 & 24.29 & 27.67 & 26.75 & 29.38 & 25.73 & 28.63 & 27.22 & 28.81 & 25.83 & 28.92 & 27.71 & 29.86 & 25.28 & 28.65 \\
 & Gemma-2-9B & 27.19 & 26.92 & 24.48 & 25.41 & 28.17 & 26.27 & 26.10 & 28.28 & 27.87 & 26.07 & 26.51 & 28.19 & 27.72 & 26.09 & 26.40 & 27.49 \\
\midrule
\multirow{3}{*}{Large} & Llama-3.1-70B & 28.40 & 31.85 & 27.64 & 29.89 & 29.78 & 31.71 & 28.85 & 31.07 & 30.06 & 31.15 & 29.16 & 30.73 & 30.12 & 31.30 & 28.75 & 31.06 \\
 & Qwen2.5-72B & 27.49 & 33.07 & 28.76 & 32.64 & 28.34 & 33.62 & 29.57 & 33.03 & 28.50 & 33.45 & 29.96 & 33.56 & 28.61 & 33.59 & 29.75 & 33.32 \\
 & Mistral-Large & \textbf{30.43} & \textbf{36.82} & \textbf{30.04} & \textbf{34.54} & \textbf{31.61} & \textbf{35.67} & \textbf{31.26} & \textbf{34.35} & \textbf{32.05} & \textbf{35.70} & \textbf{31.35} & \textbf{33.99} & \textbf{32.45} & \textbf{35.66} & \textbf{31.77} & \textbf{33.95} \\
\bottomrule
\end{tabular}}
\end{table*}

\end{document}